\documentclass[10pt,journal,compsoc]{IEEEtran}

\usepackage{graphicx}
\usepackage{cvpr}
\usepackage{amsmath}
\usepackage{amssymb}
\usepackage{multirow}
\usepackage{mathrsfs}
\usepackage{tabulary}
\usepackage{booktabs}
\usepackage{silence}
\usepackage{subcaption} % subfigure, this give a Unknown document class: https://tex.stackexchange.com/questions/567285/package-caption-warning-unknown-document-class-or-package-standard-defaults

\usepackage[bookmarks=false,colorlinks=true,linkcolor=black,citecolor=black,filecolor=black,urlcolor=black]{hyperref}
\usepackage{url}

\newcolumntype{x}[1]{>{\centering\arraybackslash}p{#1pt}}

\newlength\savewidth\newcommand\shline{\noalign{\global\savewidth\arrayrulewidth
  \global\arrayrulewidth 1pt}\hline\noalign{\global\arrayrulewidth\savewidth}}
\newcommand{\tablestyle}[2]{\setlength{\tabcolsep}{#1}\renewcommand{\arraystretch}{#2}\centering\footnotesize}
\newcommand{\parr}[1]{\left(#1\right)}

\ifCLASSOPTIONcompsoc
  \usepackage[nocompress]{cite}
\else
  \usepackage{cite}
\fi

\usepackage{url}
\begin{document}

\title{Animatable Implicit Neural Representations for Creating Realistic Avatars from Videos}

\author{Sida Peng, Zhen Xu, Junting Dong, Qianqian Wang, Shangzhan Zhang, \\
    Qing Shuai, Hujun Bao and Xiaowei Zhou

\IEEEcompsocitemizethanks{
\IEEEcompsocthanksitem S. Peng, Z. Xu, J. Dong, S. Zhang, Q. Shuai, H. Bao and X. Zhou are affiliated with the State Key Lab of CAD\&CG, the College of Computer Science, Zhejiang University, China. 
\IEEEcompsocthanksitem Q. Wang is with the College of Computer Science, Cornell University, USA.
\IEEEcompsocthanksitem Corresponding author: Xiaowei Zhou.
}
}

\markboth{Journal of \LaTeX\ Class Files,~Vol.~14, No.~8, August~2015}%
{Shell \MakeLowercase{\textit{et al.}}: Bare Demo of IEEEtran.cls for Computer Society Journals}



\IEEEtitleabstractindextext{%
% 1. Instead, we introduce neural blend weight fields to produce the deformation fields.
% 2. Based on the skeleton-driven deformation, blend weight fields are used with 3D human skeletons to generate observation-to-canonical and canonical-to-observation correspondences.
% 3. Since 3D human skeletons are more observable, they can regularize the learning of deformation fields. Moreover, the learned blend weight fields can be combined with input skeletal motions to generate new deformation fields to animate the human model.
\begin{abstract}
    This paper addresses the challenge of reconstructing an animatable human model from a multi-view video. Some recent works have proposed to decompose a non-rigidly deforming scene into a canonical neural radiance field and a set of deformation fields that map observation-space points to the canonical space, thereby enabling them to learn the dynamic scene from images. However, they represent the deformation field as translational vector field or SE(3) field, which makes the optimization highly under-constrained. Moreover, these representations cannot be explicitly controlled by input motions. Instead, we introduce a pose-driven deformation field based on the linear blend skinning algorithm, which combines the blend weight field and the 3D human skeleton to produce observation-to-canonical correspondences. Since 3D human skeletons are more observable, they can regularize the learning of the deformation field. Moreover, the pose-driven deformation field can be controlled by input skeletal motions to generate new deformation fields to animate the canonical human model. Experiments show that our approach significantly outperforms recent human modeling methods. The code is available at \href{https://zju3dv.github.io/animatable\_nerf/}{https://zju3dv.github.io/animatable\_nerf/}.
\end{abstract}

\begin{IEEEkeywords}
Human Modeling, Implicit Neural Representations, View Synthesis
\end{IEEEkeywords}}

\maketitle

\IEEEdisplaynontitleabstractindextext

% \fi
\IEEEpeerreviewmaketitle

%#################################################################################################
\begin{figure*}[t]
\centering
\includegraphics[width=1\linewidth]{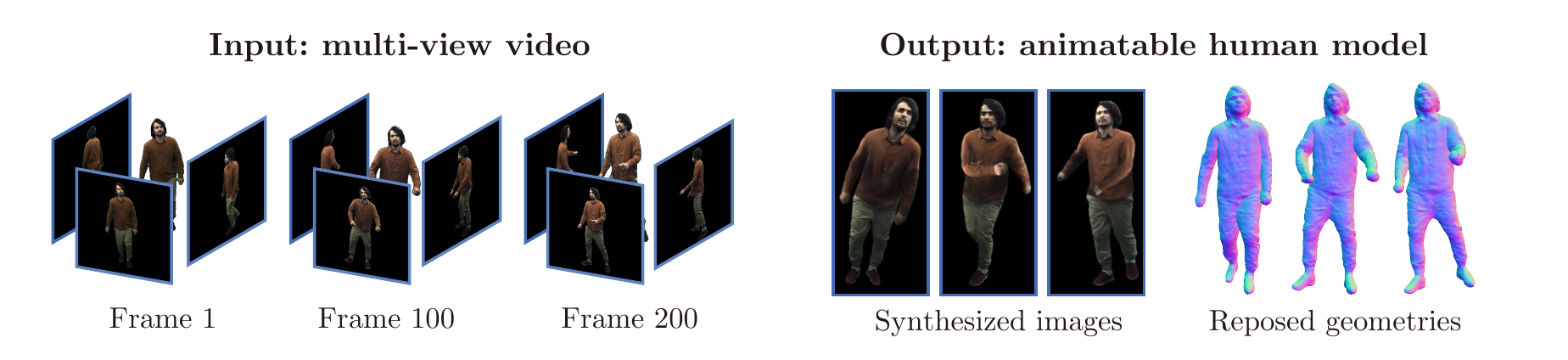}
\vspace{-1.5em}
\caption{Given a multi-view video of a performer, our method reconstructs an animatable human model, which can be used for novel view synthesis and 3D shape generation under novel human poses.}
\label{fig:teaser}
\end{figure*}
%#################################################################################################

\section{Introduction}

% the importance of human animation

Rendering animatable human characters has many applications such as free-viewpoint videos, telepresence, video games and movies. 
% Most traditional methods first reconstruct an animatable human model and then synthesize novel views using the rendering pipeline. However, creating animatable human models are expensive and time-consuming due to two factors. 
% Another version for the two sentences above:  
The core step is to reconstruct animatable human models, which tends to be expensive and time-consuming in traditional pipelines due to two factors.
First, human reconstruction generally relies on complicated hardware, such as a dense array of cameras \cite{schonberger2016structure, guo2019relightables} or depth sensors \cite{collet2015high, dou2016fusion4d}. Second, human animation requires skilled artists to manually create a skeleton suitable for the human model and carefully design skinning weights \cite{lewis2000pose} to achieve realistic animation, which takes countless human labor.

% problem statement

% This work focuses on the problem of automatically obtaining animatable humans from multi-view videos, as illustrated in Figure~\ref{fig:teaser}. This setting significantly decreases the cost of human reconstruction and animation, enabling us to create digital humans at scale. 
% Another version for the above sentences: 
In this work, we aim to reduce the cost of human reconstruction and animation, to enable the creation of digital humans at scale. Specifically, we focus on the problem of reconstructing animatable humans from multi-view videos, as illustrated in Figure~\ref{fig:teaser}. 
However, this problem is extremely challenging. There are two core questions we need to answer: how to represent animatable human models and how to learn this representation from videos?

% Our method

Recently, neural radiance fields (NeRF) \cite{mildenhall2020nerf} has proposed a representation that can be efficiently learned from images with a differentiable renderer. It represents static 3D scenes as color and density fields, which work particularly well with volume rendering techniques. To extend NeRF to handle non-rigidly deforming scenes, \cite{park2021nerfies, pumarola2020d} decompose a video into a canonical NeRF and a set of deformation fields that transform observation-space points at each video frame to the canonical space. The deformation field is represented as translational vector field \cite{pumarola2020d} or SE(3) field \cite{park2021nerfies}.
% Although they can handle dynamic scenes, there are two limitations. First, optimizing NeRF jointly with translational vector field or SE(3) field is an under-constrained problem and requires complex regularizations \cite{pumarola2020d, li2020neural}. Second, they cannot explicitly synthesize novel scenes given input motions.
Although they can handle some dynamic scenes, they are not suited for representing animatable human models due to two reasons. First, jointly optimizing NeRF with translational vector fields or SE(3) fields without motion prior is an extremely under-constrained problem \cite{pumarola2020d, li2020neural}. Second, they cannot synthesize novel scenes given input motions.

To overcome these problems, we present a novel framework that decomposes an animatable human model into a canonical neural field and a pose-driven deformation field. Specifically, the canonical neural field stores the human geometry and appearance in the canonical space. To represent the human models under observation spaces, we define a pose-driven deformation field based on the linear blend skinning algorithm \cite{lewis2000pose}, which is used to transform observation-space points to the canonical space. The proposed framework has two advantages. First, since the human pose is easy to track \cite{joo2018total}, it does not need to be jointly optimized and thus provides an effective regularization on the learning of human representations. Second, based on the linear blend skinning algorithm, we can explicitly animate the neural radiance field with input motions.

To compute the transformation from the observation space to the canonical space, the linear blend skinning algorithm requires blend weights as input. A straightforward way to obtain blend weights for any 3D point is fitting 3D points to the parametric human model \cite{loper2015smpl, pavlakos2019expressive, xu2020ghum} in the observation space. However, parametric human model generally only describes the skinned human body, and its blend weights may not accurately represent the cloth deformations. We present two ways to solve this problem. A solution is learning neural blend weight fields for input human poses, which can be optimized for each 3D point and thus can describe the cloth deformations. Inspired by \cite{tiwari21neuralgif, liu2021neural}, an alternative way is to decompose the human motion into articulated and non-rigid deformations, which introduces a pose-dependent displacement field to produce the non-rigid local deformations.

For the canonical human model, we investigate two types of neural fields to represent the human geometry and appearance. The first one is density and color fields, which are widely used in recent works on dynamic scenes \cite{li2020neural, park2021nerfies, pumarola2020d} and can be optimized from videos based on the volume rendering technique. Considering that the underlying geometry of density field tends to be noisy, we also explore the use of signed distance field (SDF) in modeling the human geometry. Compared with the density field, SDF has a well-defined surface at the zero-level set, which naturally regularizes the geometry learning.

% Experiments

We evaluate our approach on the Human3.6M \cite{ionescu2013human3}, ZJU-MoCap \cite{peng2020neural} and MonoCap \cite{habermann2020deepcap, habermann2021real, peng2022animatable} datasets. Experiments show that our approach achieve state-of-the-art performance on image synthesis. In addition, we demonstrate that using signed distance field significantly improve the reconstruction performance on SyntheticHuman \cite{peng2022animatable}.
% Compared with methods that render 2D features into images \cite{thies2019deferred, wu2020multi}, our approach naturally has a better generalization ability, as we animate human characters based on the explicit deformation model \cite{lewis2000pose}. 
% In addition, our method is able to reconstruct the 3D human shape at the canonical space and repose the geometry.

In the light of previous work, this work has the following contributions: i) We reconstruct animatable human models from videos using a novel framework, which represents a dynamic human with a canonical neural field and a pose-driven deformation field. ii) We compares different implicit neural representations for human modeling from videos. iii) Our approach demonstrates significant performance improvement on image rendering and 3D reconstruction compared to recent methods on the Human3.6M, ZJU-MoCap, MonoCap and SyntheticHuman datasets.

A preliminary version of this work appeared in ICCV 2021 \cite{peng2021animatable}. Here, the work is extended in the following ways. First, we introduce the pose-dependent displacement field as an alternative way to describe the cloth deformations. Second, the signed distance field is used to improve the learning of human geometry. We validate the effectiveness of signed distance field on 3D reconstruction on a synthetic dataset. Third, we compare with more recent methods and additionally perform experiments on a monocular dataset \cite{habermann2020deepcap, habermann2021real}. More ablation studies are added to evaluate the proposed components. The code has been publicly available at \href{https://github.com/zju3dv/animatable\_nerf}{https://github.com/zju3dv/animatable\_nerf}, which has received more than 220 stars.

%#################################################################################################
\begin{figure*}[t]
\centering
\includegraphics[width=1\linewidth]{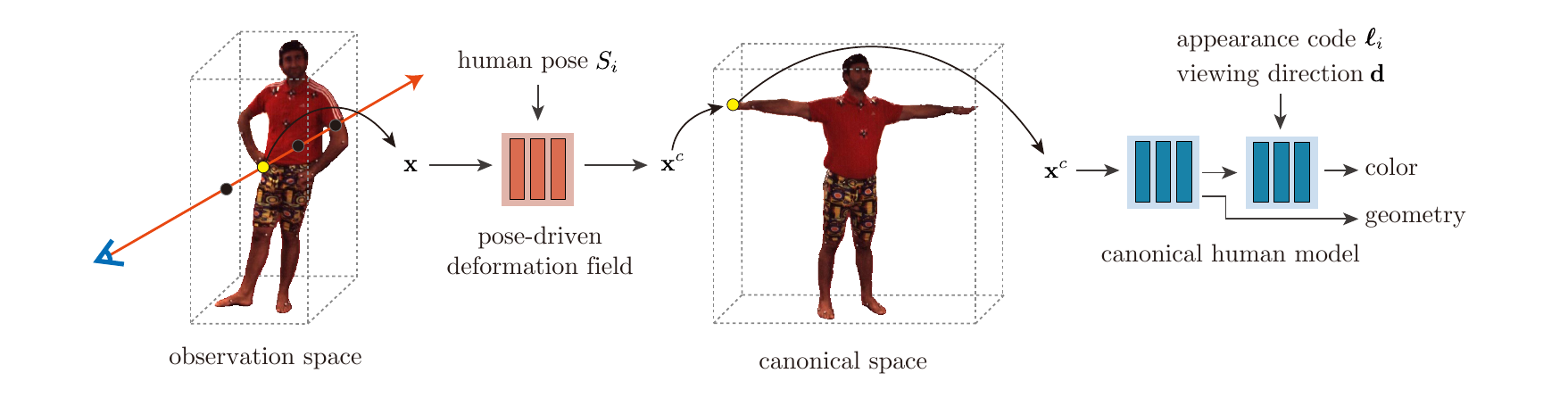}
\vspace{-2em}
\caption{\textbf{Overview of the animatable implicit neural representation.} Given a query point $\mathbf{x}$ in the observation space and human pose $S_i$ at frame $i$, we transform the observation-space point to the canonical space using the pose-driven deformation field. Taking the transformed point $\mathbf{x}^c$, viewing direction $\mathbf{d}$, and appearance code $\boldsymbol{\ell}_i$ as inputs, the canonical human model predicts the geometry and color.}
\label{fig:pipeline}
\end{figure*}
%#################################################################################################

\section{Related work}

% 1. Complicated hardware methods
% 2. Sparse camera views: nerf, dvr, idr
% 3. Single-view

\noindent \textbf{Human reconstruction.} Modeling human characters is the first step of traditional animation pipelines. To achieve high-quality reconstruction, most methods rely on complicated hardware \cite{collet2015high, dou2016fusion4d, su2020robustfusion, debevec2000acquiring, guo2019relightables}. Recently, some works \cite{sitzmann2019scene, niemeyer2020differentiable, mildenhall2020nerf, yariv2020multiview} have attempted to learn 3D representations from images with differentiable renderers, which reduces the number of input camera views and achieves impressive reconstruction results. However, they have difficulty in recovering reasonable 3D human shapes when the camera views are too sparse, as shown in \cite{peng2020neural}. Instead of optimizing the network parameters per scene, \cite{natsume2019siclope, saito2019pifu, zheng2019deephuman, saito2020pifuhd} utilize networks to learn human shape priors from ground-truth 3D data, allowing them to reconstruct human shapes from even a single image.

% 1. LBS: a common approach to animate human models
% 2. SMPL: SMPL-based methods, the geometry are too coarse
% 3. People-Snapshot, TexShape: refine SMPL
% 4. photo-wakeup: image space, does not allow novel view synthesis
% 5. Arch, IP-Net

\noindent \textbf{Human animation.} Skeletal animation \cite{lewis2000pose, kavan2007skinning} is a common approach to animate human models. It first creates a scale-appropriate skeleton for the human mesh and then assigns each mesh vertex a blend weight that describes how the vertex position deforms with the skeleton. Skinned multi-person linear model (SMPL) \cite{loper2015smpl} learns a skeleton regressor and blend weights from a large amount of ground-truth 3D meshes. Based on SMPL, some works \cite{pavlakos2018learning, kanazawa2018end, kolotouros2019cmr, jiang2020mpshape, dong2020motion} reconstruct an animated human mesh from sparse camera views. However, SMPL only describes the naked human body and thus cannot be directly used to render photorealistic images. To overcome this problem, \cite{alldieck2018video, alldieck2019learning, alldieck2019tex2shape} apply vertex displacements to the SMPL model to capture the human clothing and hair. \cite{weng2019photo} proposes a 2D warping method to deform the SMPL model to fit the input image. Recent implicit function-based methods \cite{park2019deepsdf, mescheder2019occupancy, chibane2020implicit} have exhibited state-of-the-art reconstruction quality. \cite{huang2020arch, bhatnagar2020ipnet} combine implicit function learning with the SMPL model to obtain detailed animatable human models. \cite{deng2020nasa} combines a set of local implicit functions with human skeletons to represent dynamic humans. \cite{yang2021s3, dong2022pina, chen2021snarf, saito2021scanimate, mihajlovic2021leap} propose to animate implicit neural representations with the linear blend skinning algorithm.
% However, these methods all need the supervision of 3D ground-truth data.

% 1. Motion retargeting: not robust to significant view changes.
% 2. 3D representation.
% 3. 3D-to-2D rendering.

\noindent \textbf{Neural rendering.} To reduce the requirement for the reconstruction quality, some methods \cite{shysheya2019textured, thies2019deferred, liu2020neural, wu2020multi, kwon2020rotationally} improve the rendering pipeline with neural networks. Based on the advances in image-to-image translation techniques \cite{isola2017image}, \cite{ma2017pose, chan2019everybody, men2020controllable} train a network to map 2D skeleton images to target rendering results. Although these methods can synthesize photorealistic images under novel human poses, they have difficulty in rendering novel views. To improve the performance of novel view synthesis, \cite{shysheya2019textured, thies2019deferred, wu2020multi, aliev2020neural, prokudin2021smplpix, yoon2021pose, raj2020anr} introduce 3D representations into the rendering pipeline. \cite{thies2019deferred} establishes neural texture maps and uses UV maps to obtain feature maps in the image space, which is then interpreted into images with a neural renderer. \cite{wu2020multi, aliev2020neural} reconstruct a point cloud from input images and learn a 3D feature for each point. Then, they project 3D features into a 2D feature map and employ a network to render images. However, 2D convolutional networks have difficulty in rendering inter-view consistent images, as shown in \cite{sitzmann2019scene}.

% 1. 为了解决view inconsistency的问题，一些工作在3D space预测color，然后accumulate them into 2D images。
% 2. 这些工作\cite{yao2021dd, xu2021h, kwon2021neural, , peng2020neural}预测三维人体模型。
% 3. \cite{peng2020neural}的做法。
% 4. 这是个很快发展的领域，有很多concurrent works \cite{chen2021geometry, zhao2021humannerf, jiang2022selfrecon, , noguchi2021neural, raj2022dracon, chen2022uv, weng2022humannerf, hu2021hvtr, zheng2022structured}。

To solve this problem, \cite{lombardi2019neural, niemeyer2020differentiable, mildenhall2020nerf, li2020crowd, liu2020nsvf, suo2021neuralhumanfvv} interpret features into colors in 3D space and then accumulate them into 2D images. In the field of human modeling, \cite{yao2021dd, xu2021h, kwon2021neural, peng2020neural, noguchi2021neural} represent 3D human models as implicit neural representations and optimize network parameters from images with differentiable volume rendering. \cite{peng2020neural} combines neural radiance field with the SMPL model, allowing it to handle dynamic humans and synthesize photorealistic novel views from very sparse camera views. Reconstructing 3D humans from videos is a fast growing field, and there are many concurrent works \cite{chen2021geometry, zhao2021humannerf, jiang2022selfrecon, raj2022dracon, chen2022uv, weng2022humannerf, hu2021hvtr, zheng2022structured}. Similar to \cite{peng2021animatable}, \cite{weng2022humannerf, zheng2022structured, jiang2022selfrecon, raj2022dracon} leverage the LBS model to establish observation-to-canonical correspondences, which enables them to aggregate temporal observations in the input video.

\section{Method}

% Overview

Given a multi-view video of a performer, our task is to reconstruct an animatable human model that can be used to synthesize free-viewpoint videos of the performer under novel human poses. The cameras are synchronized and calibrated. For each frame, we assume the 3D human skeleton is given, which can be obtained with marker-based or marker-less pose estimation systems \cite{ionescu2013human3, joo2018total}. For each image, \cite{gong2018instance} is used to extract the foreground human mask, and the values of the background image pixels are set as zero.

The overview of our approach is shown in Figure~\ref{fig:pipeline}. We decompose a non-rigidly deforming human body into a canonical human model represented by a neural field (Section~\ref{section:nerf}) and a pose-driven deformation field (Section~\ref{section:nsf}) that is used to establish observation-to-canonical correspondences. Then we discuss how to learn the representation on the multi-view video (Section~\ref{section:optim}). Based on the learned pose-driven deformation field, we are able to explicitly animate the canonical human model (Section~\ref{section:anerf}).

\subsection{Representing videos with neural fields}

\label{section:nerf}

Neural radiance field (NeRF) represents a static scene as a continuous representation. For any 3D point, it takes a spatial position $\mathbf{x}$ and viewing direction $\mathbf{d}$ as input to a neural network and outputs the density $\sigma$ and color $\mathbf{c}$.

Inspired by \cite{park2021nerfies, pumarola2020d}, we extend NeRF to represent the dynamic human body by introducing a pose-driven deformation field, as shown in Figure~\ref{fig:pipeline}. Specifically, for a video frame $i$ with a 3D human pose $S_i$, we define a deformation field $T$ that transforms observation-space points to the canonical space. Given the canonical-frame geometry model $F_{g}$, the geometry model at frame $i$ can be thus defined as:
\begin{equation}
    (\sigma_i(\mathbf{x}), \mathbf{z}_i(\mathbf{x})) = F_{g}(\gamma_\mathbf{x}(T(\mathbf{x}, S_i))),
    \label{eq:sigma}
\end{equation}
where $\mathbf{z}_i(\mathbf{x})$ is the shape feature in the original NeRF, and $\gamma_\mathbf{x}$ is the positional encoding \cite{mildenhall2020nerf} for spatial location. The density model $F_{g}$ is implemented as an MLP network with nine fully-connected layers.

When predicting the color, we define a per-frame latent code $\boldsymbol{\ell}_i$ to encode the state of the human appearance in frame $i$. Similarly, with the canonical-frame color model $F_{\mathbf{c}}$, the color model at frame $i$ can be defined as:
\begin{equation}
    \mathbf{c}_i(\mathbf{x}) = F_{\mathbf{c}}(\mathbf{z}_i(\mathbf{x}), \gamma_\mathbf{d}(\mathbf{d}), \boldsymbol{\ell}_i),
    \label{eq:color}
\end{equation}
where $\gamma_\mathbf{d}$ is the positional encoding for viewing direction, and $F_{\mathbf{c}}$ represents an MLP network with five layers.

\subsubsection{Signed distance fields}

An alternative way to model the human geometry is using the signed distance field. In contrast to density field, signed distance field has a well-defined surface at the zero-level set, which facilitates more direct regularization on the geometry learning and generally achieves better reconstruction performance \cite{wang2021neus, yariv2021volume}. In practice, since signed distance is a scalar, we can predict the signed distance $s$ for each 3D point $\mathbf{x}$ using the same geometry network $F_g: \mathbf{x} \rightarrow s$. More implementation details can be found in the supplementary material. We apply Eikonal constraint \cite{gropp2020implicit} to enforce the network prediction to conform with the property of signed distance field, which will be described in Section~\ref{section:optim}.

\subsection{Pose-driven deformation fields}

\label{section:nsf}

Given the canonical human model, we use a pose-driven deformation field to obtain the human model under a particular human pose. There are several ways to represent the deformation field, such as translational vector field \cite{pumarola2020d, li2020neural} and SE(3) field \cite{park2021nerfies}. However, as discussed in \cite{park2021nerfies, li2020neural}, optimizing a neural field along with a deformation field is an ill-posed problem that is prone to local optima. Moreover, their representations cannot be explicitly driven by novel motion sequences. Considering that we aim to represent dynamic humans, it is natural to leverage the human priors to build the deformation field, which helps us to solve the under-constrained problem. Specifically, we construct the deformation field based on the 3D human pose and the pose-driven deformation framework \cite{lewis2000pose}. 

The human pose defines $K$ parts, which produce $K$ transformation matrices $\{G_k\} \in SE(3)$. The detailed derivation is listed in the supplementary material. In the linear blend skinning algorithm \cite{lewis2000pose}, a canonical-space point $\mathbf{v}$ is transformed to the observation space using
\begin{equation}
    \mathbf{v}' = \left( \sum_{k=1}^K w(\mathbf{v})_k G_k \right) \mathbf{v},
    \label{eq:skinning}
\end{equation}
where $w(\mathbf{v})_k$ is the blend weight of $k$-th part. Similarly, for an observation-space point $\mathbf{x}$, if we know its corresponding blend weights, we are able to transform it to the canonical space using
\begin{equation}
    \mathbf{x}' = \left( \sum_{k=1}^K w^o(\mathbf{x})_k G_k \right)^{-1} \mathbf{x},
    \label{eq:inverse}
\end{equation}
where $w^o(\mathbf{x})$ is the blend weight function defined in the observation space. To obtain the blend weight field, a straightforward way is to calculate the blend weights for each 3D point based on the parametric human model \cite{loper2015smpl, romero2017embodied, pavlakos2019expressive, xu2020ghum}. Without loss of generality, we adopt SMPL \cite{loper2015smpl} as the parametric model and compute the blend weight field using the strategy in \cite{huang2020arch, bhatnagar2020loopreg}. Specifically, for any 3D point, we first find the closest surface point on the SMPL mesh. Then, the target blend weight is obtained by retrieving the blend weights of the nearest vertex on the SMPL model. 

However, parametric model cannot describe personalized human details, which makes the deformation inaccurate and could degrade the performance, as demonstrated by the experimental results in Section~\ref{sec:ablation}. To solve this problem, we exploit two ways. One is learning neural blend weight fields that are optimized for each 3D point to produce more accurate deformation fields. The second way is introducing a pose-dependent displacement field to compensate for the inaccurate deformation.

\subsubsection{Neural blend weight fields}

For any 3D point, we first retrieve an initial blend weight from the parametric body model and then use a network to learn a residual vector, resulting in the neural blend weight field. In practice, the residual vector fields for all training video frames are implemented using a single MLP network $F_{\Delta \mathbf{w}}: (\mathbf{x}, \boldsymbol{\psi}_i) \rightarrow \Delta \mathbf{w}_i$, where $\boldsymbol{\psi}_i$ is a per-frame learned latent code and $\Delta \mathbf{w}_i$ is a vector $\in \mathbb{R}^{K}$. The neural blend weight field at frame $i$ is defined as:
\begin{equation}
    \mathbf{w}_i(\mathbf{x}) = \text{norm}(F_{\Delta \mathbf{w}}(\mathbf{x}, \boldsymbol{\psi}_i) + \mathbf{w}^{\text{s}}(\mathbf{x}, S_i)),
    \label{eq:lbw}
\end{equation}
where $\mathbf{w}^{\text{s}}$ is the initial blend weights that are computed based on the parametric body model under the human pose $S_i$, and we define $\text{norm}(\mathbf{w})=\mathbf{w}/ \sum w_i$. 

To animate the template human model, we additionally learn a neural blend weight field $\mathbf{w}^{\text{can}}$ at the canonical space. The SMPL blend weight field $\mathbf{w}^{\text{s}}$ is calculated using the canonical SMPL model, and $F_{\Delta \mathbf{w}}$ is conditioned on an additional latent code $\boldsymbol{\psi^{\text{can}}}$. We utilize the consistency between blend weights to optimize the neural blend weight field $\mathbf{w}^{\text{can}}$, which is described in Section~\ref{section:optim}. During animation, the canonical blend weight field is used to compute blend weights for observation-space coordinates under unseen human poses, which is described in Section~\ref{section:anerf}.

Instead of calculating blend weights of novel human poses from the canonical blend weight field $\mathbf{w}^{\text{can}}$, an alternative method is to define the pose-dependent residual vector field $F'_{\Delta \mathbf{w}}: (\mathbf{x}, S) \rightarrow \Delta \mathbf{w}$, which enables us to directly predict the blend weight field under human pose $S$ using Equation~\eqref{eq:lbw}. However, the input coordinate $\mathbf{x}$ for $F'_{\Delta \mathbf{w}}$ is in the observation space, whose value could vary significantly with human pose, making the residual vector field $F'_{\Delta \mathbf{w}}$ difficult to generalize to unseen poses.

\subsubsection{Pose-dependent displacement fields}

Another way to improve the SMPL-based deformation is introducing the pose-dependent displacement field \cite{tiwari21neuralgif, liu2021neural}. We decompose the human motion into articulated and non-rigid deformations, which is represented by the LBS model \cite{lewis2000pose, peng2021animatable} and a neural displacement field, respectively.

Specifically, for an observation-space point $\mathbf{x}$ at frame $i$, we first obtain its blend weights from the SMPL model and warp it to the canonical space using Equation~\eqref{eq:inverse}, resulting in the transformed point $\mathbf{x}'$. Then, we use a displacement field to deform the point $\mathbf{x}'$ to the surface. Denote the displacement field as $F_{\Delta \mathbf{x}}: (\mathbf{x}, S_i) \rightarrow \Delta \mathbf{x}_i$, where $S_i$ is the 3D human pose at frame $i$. The final point is $\mathbf{x}' + F_{\Delta \mathbf{x}}(\mathbf{x}', S_i)$. The displacement field is implemented as an MLP network. 

The experiments show that the pose-dependent displacement field $F_{\Delta \mathbf{x}}$ can generalize to unseen human poses. A plausible reason is that $F_{\Delta \mathbf{x}}$ takes the canonical-space coordinate as input, whose value is similar across different human poses. Compared with neural blend weight field, the displacement field is more convenient to use, as neural blend weight field additionally requires us to calculate blend weights under novel poses during animation. More details can be found in Section~\ref{section:anerf}.

\subsection{Training}

\label{section:optim}

Given the animatable neural field defined in Equations \eqref{eq:sigma} and \eqref{eq:color}, we can use volume rendering techniques \cite{kajiya1986rendering, mildenhall2020nerf} to synthesize images of particular viewpoints for each video frame $i$. The near and far bounds of volume rendering are estimated by computing the 3D boxes that bound the SMPL meshes. When the density field is used to model the human geometry, we directly use the rendering strategy in \cite{mildenhall2020nerf} to render pixel colors. The parameters of canonical neural field and pose-driven deformation field are jointly optimized over the multi-view video by minimizing the difference between the rendered pixel color $\tilde{\mathbf{C}}_i(r)$ and the observed pixel color $\mathbf{C}_i(r)$:
\begin{equation}
    L_{\text{rgb}} = \sum_{r \in \mathcal{R}} \|\tilde{\mathbf{C}}_i(\mathbf{r}) - \mathbf{C}_i(\mathbf{r}) \|_2,
\end{equation}
where $\mathcal{R}$ is the set of rays passing through image pixels.

\textbf{Additional losses for learning signed distance field.} When the human geometry is represented as signed distance field (SDF), we first convert predicted signed distances into densities using the strategy in VolSDF \cite{yariv2021volume} and then synthesize pixel colors through the volume rendering. In addition, the mask loss and the Eikonal term \cite{gropp2020implicit} are used for learning the SDF. To supervise the SDF with the mask, we find the minimal SDF value $s^{\mathbf{r}}_i$ of sampled points along the ray $\mathbf{r}$ and apply the binary cross entropy loss BCE:
\begin{equation}
    L_{\text{mask}} = \sum_{r \in \mathcal{R}} \text{BCE}(\text{sigmoid}(-\rho s^{\mathbf{r}}_i), M_i(\mathbf{r})),
\end{equation}
where $M_i(\mathbf{r}) \in \{0, 1\}$ is the ground-truth mask value. Similar to \cite{yariv2020multiview}, we set $\rho$ as 50 and multiply it by 2 every 10000 iterations. The number of multiplications is up to 5. We sample a set of points $\mathcal{X}_i$ in the observation space and apply the Eikonal term on these sampled points:
\begin{equation}
    L_{\text{E}} = \sum_{\mathbf{x} \in \mathcal{X}_i} (\| \nabla F_g(T(\mathbf{x}, S_i)) \|_2 - 1)^2.
\end{equation}

\textbf{Additional losses for learning deformation field.} When the neural blend weight field is used to represent the deformation field, we introduce a consistency loss to learn the neural blend weight field $\mathbf{w}^{\text{can}}$ at the canonical space. As shown by Equations \eqref{eq:skinning} and \eqref{eq:inverse}, two corresponding points at canonical and observation spaces should have the same blend weights. For an observation-space point $\mathbf{x}$ at frame $i$, we map it to the canonical-space point $T_i(\mathbf{x})$ using Equation \eqref{eq:inverse}. The consistency loss is defined as:
\begin{equation}
    L_{\text{nsf}} = \sum_{\mathbf{x} \in \mathcal{X}'_i} \| \mathbf{w}_i(\mathbf{x}) - \mathbf{w}^{\text{can}}(T_i(\mathbf{x})) \|_1,
\end{equation}
where $\mathcal{X}'_i$ is the set of 3D points sampled within the 3D human bounding box at frame $i$.

When the displacement field is used to produce the deformation field, we apply the regularization to predicted displacements, which is defined as:
\begin{equation}
    L_{\Delta \mathbf{x}} = \sum_{\mathbf{x} \in \mathcal{X}''_i} \| F_{\Delta \mathbf{x}}(\mathbf{x}, S_i) \|_2,
\end{equation}
where $\mathcal{X}''_i$ is the set of 3D points sampled in the canonical space. The coefficient weights of losses described above are specified in the supplementary material.

\subsection{Animation}

\label{section:anerf}

After training, we can use the pose-driven deformation field to animate the canonical human model. Given the novel human pose $S^{\text{new}}$, the pose-driven deformation field warps observation-space points to the canonical space, which are then fed into the geometry and color models. Note that the color model is conditioned on a latent code, which is not defined for unseen human pose. To solve this problem, we select the latent code of the training human pose that is nearest to the novel human pose $S^{\text{new}}$.

When the deformation field is represented as neural blend weight field, we need to additionally optimize blend weights under novel poses based on the canonical blend weight field. Specifically, for the novel human pose $S^{\text{new}}$, our method first computes the SMPL blend weight field $\mathbf{w}^{\text{s}}$. Then, the neural blend weight field $\mathbf{w}^{\text{new}}$ for the novel human pose is defined as:
\begin{equation}
    \resizebox{.8\hsize}{!}{$\mathbf{w}^{\text{new}}(\mathbf{x}, \boldsymbol{\psi}^{\text{new}}) = \text{norm}(F_{\Delta \mathbf{w}}(\mathbf{x}, \boldsymbol{\psi}^{\text{new}})  +  \mathbf{w}^{\text{s}}(\mathbf{x}, S^{\text{new}})),$}
\end{equation}
where the $F_{\Delta \mathbf{w}}$ is conditioned on a new latent code $\boldsymbol{\psi}^{\text{new}}$. Based on the $\mathbf{w}^{\text{new}}$ and Equation \eqref{eq:inverse}, we can generate the deformation field $T^{\text{new}}$ for the novel human pose. The parameters of $\boldsymbol{\psi}^{\text{new}}$ are optimized using
\begin{equation}
    L_{\text{new}} = \sum_{\mathbf{x} \in \mathcal{X}^{\text{new}}} \| \mathbf{w}^{\text{new}}(\mathbf{x}) - \mathbf{w}^{\text{can}}(T^{\text{new}}(\mathbf{x})) \|_1,
    \label{eq:lnew}
\end{equation}
where $\mathcal{X}^{\text{new}}$ is the set of 3D points sampled within the human box under the novel human pose. Note that we fix the parameters of $\mathbf{w}^{\text{can}}$ during training.  In practice, we train neural skinning fields under multiple novel human poses simultaneously. This is implemented by conditioning $F_{\Delta \mathbf{w}}$ on multiple latent codes. With the deformation field $T^{\text{new}}$, our method uses Equations \eqref{eq:sigma} and \eqref{eq:color} to produce the human model under the novel human pose. 

When the pose-dependent displacement field $F_{\Delta \mathbf{x}}$ is used, we can establish the observation-to-canonical correspondences by first warping observation-space points to the canonical space and then deforming them using $F_{\Delta \mathbf{x}}$. In contrast to neural blend weight field, the pose-dependent displacement field does not require the additional optimization during animation, which is easier to use.

% To suppress the noise caused by inaccurate correspondences, the points' densities are set to zero if their distances to the SMPL mesh surface are bigger than a threshold $\tau$. We set $\tau$ to $5 cm$ in all experiments.

% \paragraph{3D shape generation.} In addition to synthesizing images under novel human poses, our approach can also explicitly animate a reconstructed human mesh, similar to the traditional animation methods. In particular, we first discretize the human bounding box at the canonical space with a voxel size of $5mm \times 5mm \times 5mm$ and evaluate the volume densities for all voxels, which are used to extract the human mesh with the Marching Cubes algorithm \cite{lorensen1987marching}. Then, blend weights of mesh vertices are inferred from the neural blend weight field $\mathbf{w}^{\text{can}}$. Finally, given a novel human pose, we use equation \eqref{eq:skinning} to transform each vertex, resulting in a deformed mesh under the target pose. The reconstruction results are presented in the supplementary material.

\section{Implementation details}

The networks of our canonical human model $F_{g}$ and $F_{\mathbf{c}}$ closely follow the original IDR \cite{yariv2020multiview}. The networks of $F_{\Delta \mathbf{w}}$ and $F_{\Delta \mathbf{x}}$ consist of nine fully connected layers. More details of network architectures are described in the supplementary material. The appearance code $\boldsymbol{\ell}_i$ and blend weight field code $\boldsymbol{\psi}_i$ both have dimensions of 128.

\textbf{Training.} The Adam optimizer \cite{kingma2014adam} is adopted for the training. The learning rate starts from $5e^{-4}$ and decays exponentially to $5e^{-5}$ along the optimization. The training is conducted on a 2080 Ti GPU. For a three-view video of 300 frames, the training takes around 200k iterations to converge. During animation, we use the same optimizer and learning rate scheduler to optimize the neural blend weight field. For 200 novel human poses, the optimization takes around 10k iterations to converge. More details about training can be found in the supplementary material.

%#################################################################################################
\begin{figure*}[t]
\centering
\includegraphics[width=0.95\linewidth]{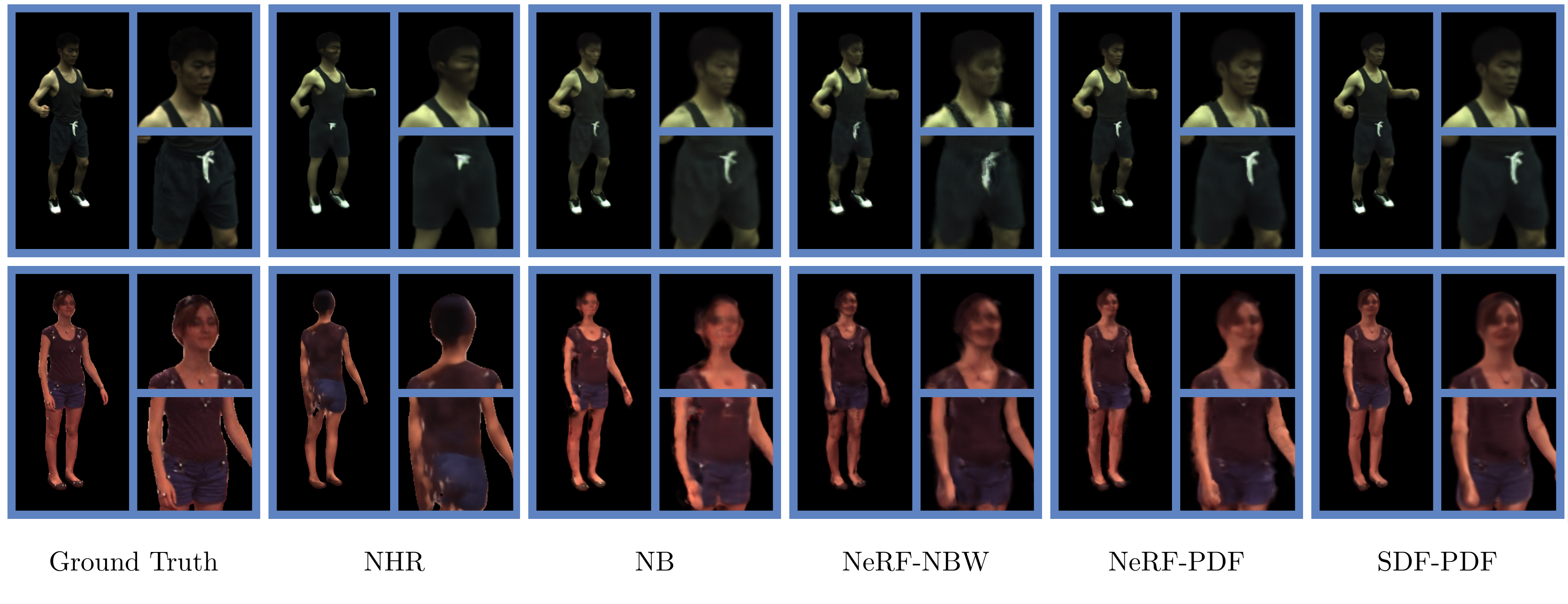}
\vspace{-1em}
\caption{\textbf{Qualitative results of novel view synthesis under training human poses on the ZJU-MoCap and Human3.6M dataset.} The first and second rows present results from the ZJU-MoCap dataset and the Human3.6M dataset respectively. ``NB" means Neural Body. ``NeRF-NBW", ``NeRF-PDF" and ``SDF-PDF" are our animatable human representations. \cite{wu2020multi} have difficulty in controlling the viewpoint and seem to overfit training views. Compared with them, our proposed representations accurately render the target view.}
\vspace{-1em}
\label{fig:training_pose_novel_view}
\end{figure*}
%#################################################################################################

\section{Experiments}

We evaluate three types of animatable human representations: a) NeRF with neural blend weight field (NeRF-NBW), b) NeRF with pose-dependent displacement field (NeRF-PDF), and c) signed distance and color fields with pose-dependent displacement field (SDF-PDF). Here the signed distance and color fields with neural blend weight field is not evaluated, because we found that this representation is prone to local minima during optimization.

\subsection{Dataset and metrics}

We describe the datasets and metrics used in experiments. More details of datasets and metrics can be found in the supplementary material.

\noindent \textbf{Human3.6M \cite{ionescu2013human3}} records multi-view videos with 4 cameras and collects human poses using the marker-based motion capture system. It includes multiple subjects performing complex actions. We select representative actions, split the videos into training and test frames, and perform experiments on subjects S1, S5, S6, S7, S8, S9, and S11. Three cameras are used for training and the remaining camera is selected for test. We use \cite{joo2018total} to obtain the SMPL parameters from the 3D human poses and apply \cite{gong2018instance} to segment foreground humans.

\noindent \textbf{MonoCap} is created by \cite{peng2022animatable}, which consists of two videos from DeepCap dataset \cite{habermann2020deepcap} and two videos from DynaCap dataset \cite{habermann2021real}, which are captured by dense camera views and provide the human masks and 3D human poses. We use one camera view for training and select ten uniformly distributed cameras for test. We select a clip of each video to perform experiments. Each clip has 300 frames for training and 300 frames for evaluating novel pose synthesis.

% 在单目数据集上blend weight field效果差一些，可能是因为blend weight比较难学

\noindent \textbf{ZJU-MoCap \cite{peng2020neural}} records 9 multi-view videos with 21 cameras and collects human poses using the marker-less motion capture system. Following the experimental protocol in \cite{peng2020neural}, we select four uniformly distributed cameras as training input and test on the remaining cameras. To further explore the capability of our method that aggregates the temporal observations, we additionally train models on the first camera view and test them on the remaining cameras.

\noindent \textbf{SyntheticHuman} is a synthetic dataset created by \cite{peng2022animatable}, which contains 7 animated 3D characters. 4 human characters perform rotation while holding A-pose, which are rendered into monocular videos. Another 3 human characters perform random actions, which are rendered with four cameras. All video frames and camera views are used for training. This dataset is only used to evaluate the performance on 3D reconstruction.

\begin{table}
\begin{center}
\scalebox{0.92}{
\tablestyle{4pt}{1.2}
\begin{tabular}{l|x{32}x{32}|x{32}x{32}}
& \multicolumn{2}{c|}{Training poses} & \multicolumn{2}{c}{Novel poses} \\[.1em]
% \shline
& PSNR$\uparrow$ & SSIM$\uparrow$ & PSNR$\uparrow$ & SSIM$\uparrow$ \\
\hline
D-NeRF \cite{pumarola2020d}  & 20.13          & 0.807          & -              & -              \\
NB \cite{peng2020neural}     & 23.31          & 0.903          & 22.74          & 0.885          \\
NHR \cite{wu2020multi}       & 20.93          & 0.866          & 20.47          & 0.857          \\
A-NeRF \cite{su2021anerf}    & 23.95          & 0.906          & 22.54          & 0.882          \\
NeRF-NBW                     & 24.46          & 0.901          & 23.65          & 0.890          \\
NeRF-PDF                     & \textbf{24.75} & 0.907          & \textbf{23.78} & 0.892          \\
SDF-PDF                      & 24.71          & \textbf{0.914} & 23.67          & \textbf{0.899} \\
\end{tabular}
}
\end{center}
\vspace{-1.3em}
\caption{\textbf{Results of novel view synthesis of training poses and novel poses on Human3.6M dataset in terms of PSNR and
SSIM (higher is better).} ``NB" means Neural Body. ``NeRF-NBW", ``NeRF-PDF" and ``SDF-PDF" are our animatable human representations. All methods are trained on three views and tested on one view.}
\vspace{-1.2em}
\label{table:h36m_result}
\end{table}

%\vspace{1em}

% In addition, we apply \cite{gong2018instance} to segment foreground humans. More details of training and test data can be found in the supplementary material.

\noindent \textbf{Metrics.} For 3D reconstruction, we follow \cite{saito2019pifu} to use two metrics: point-to-surface Euclidean distance (P2S) and Chamfer distance (CD). Units for the two metrics are in cm. For image synthesis, we follow \cite{mildenhall2020nerf} to evaluate our method using two metrics: peak signal-to-noise ratio (PSNR) and structural similarity index (SSIM).

%#################################################################################################
\begin{figure*}[t]
\centering
\includegraphics[width=0.95\linewidth]{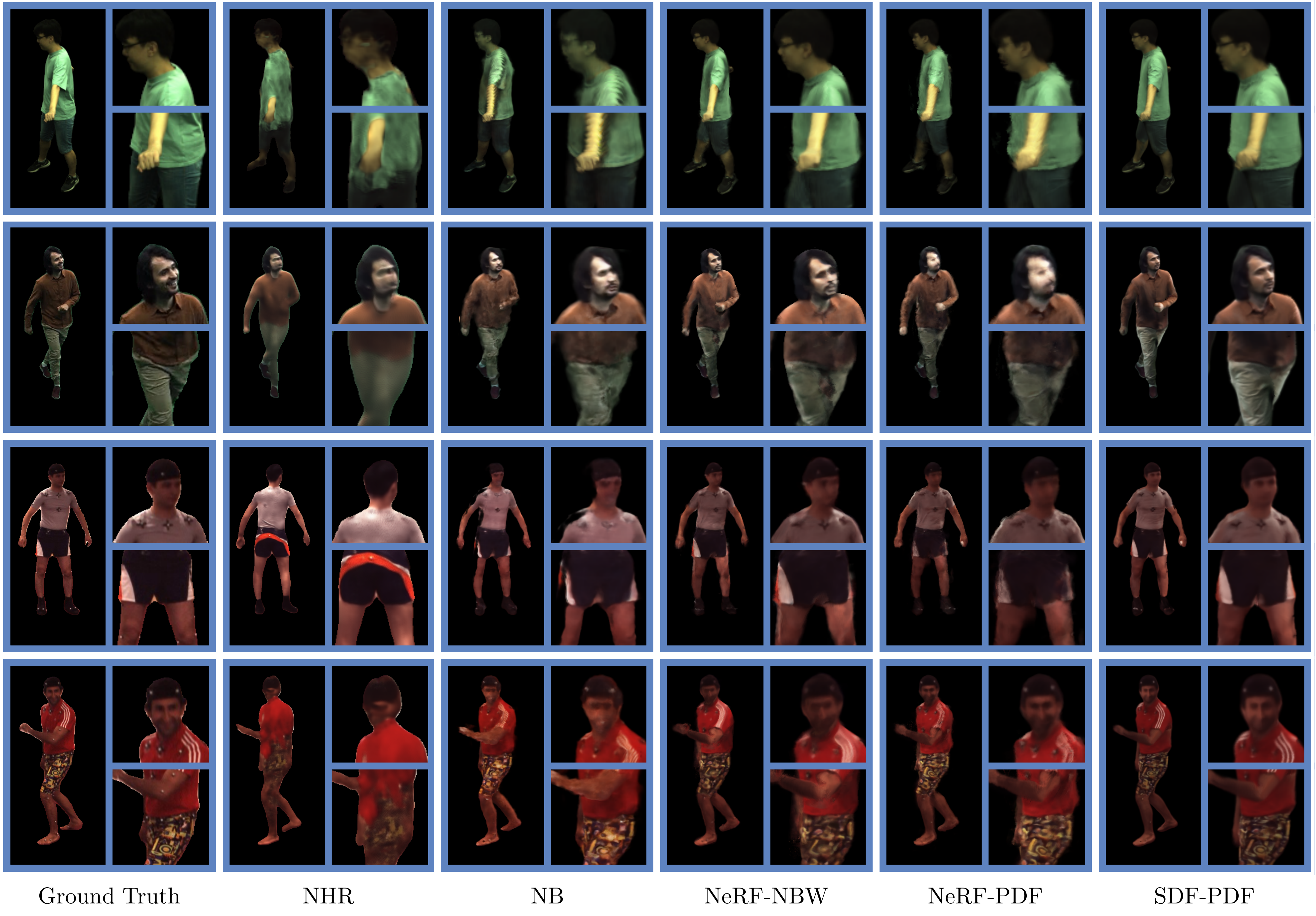}
% \vspace{-1em}
\caption{\textbf{Qualitative results of novel pose synthesis on the ZJU-MoCap, MonoCap and Human3.6M datasets.} Figures on the first row are results from the ZJU-MoCap dataset. The following row come from the MonoCap dataset. The last two rows are from the Human3.6M dataset. For complex human poses, \cite{wu2020multi, peng2020neural} tend to generate distorted rendering results. In contrast to them, our method has a better generalization ability.}
% \vspace{-1em}
\label{fig:novel_pose_novel_view}
\end{figure*}
%#################################################################################################

\subsection{Performance on image synthesis}

Our method is compared with \cite{wu2020multi, peng2020neural, pumarola2020d, su2021anerf} that train a separate network for each video. We do not compare with \cite{liu2021neural, xu2021h}, because they do not release the source code. Detailed descriptions of the baseline methods are presented in the supplementary material.

\textbf{Results on the Human3.6M dataset.} Table~\ref{table:h36m_result} compares our method with \cite{wu2020multi, peng2020neural, pumarola2020d, su2021anerf} on image synthesis. \cite{wu2020multi} extracts point descriptors from the SMPL point cloud and renders 2D feature maps, which are then interpreted into an image using a 2D CNN. \cite{peng2020neural} anchors a set of latent codes to the SMPL model and regresses a neural radiance field from these latent codes. \cite{su2021anerf} constructs the skeleton-relative encoding to predict the human model under a given human pose. On novel view synthesis of training and novel poses, all three of our proposed representations outperform the baseline methods.

In Figure~\ref{fig:training_pose_novel_view} and \ref{fig:novel_pose_novel_view}, we present qualitative results of our method and baseline methods on novel view synthesis of training and novel poses. Our method produces photo-realistic rendering results and outperforms baseline methods. We can see that \cite{wu2020multi} has difficulty in controlling the rendering viewpoint and tend to synthesize contents of training views. As shown in the third person of Figure~\ref{fig:novel_pose_novel_view}, they render the human back that is seen during training. Although \cite{peng2020neural} synthesizes high-quality images on training poses, it struggles to give reasonable rendering results on novel human poses. In contrast, our method explicitly animates the canonical human model using a pose-driven deformation field, which is similar to the classical graphics pipeline and has better controllability on the image generation process than CNN-based methods. The supplementary material presents more qualitative results.

\begin{table}
\begin{center}
\scalebox{0.92}{
\tablestyle{4pt}{1.2}
\begin{tabular}{l|x{32}x{32}|x{32}x{32}}
& \multicolumn{2}{c|}{Training poses} & \multicolumn{2}{c}{Novel poses} \\[.1em]
% \shline
& PSNR$\uparrow$ & SSIM$\uparrow$ & PSNR$\uparrow$ & SSIM$\uparrow$ \\
\hline
NB \cite{peng2020neural}     & 21.76          & 0.872          & 20.83          & 0.854          \\
NHR \cite{wu2020multi}       & 21.29          & 0.875          & 20.45          & 0.866          \\
A-NeRF \cite{su2021anerf}    & 20.52          & 0.845          & 19.53          & 0.828          \\
NeRF-NBW                     & 21.47          & 0.868          & 20.66          & 0.860          \\
NeRF-PDF                     & \textbf{22.34} & 0.883          & \textbf{21.19} & 0.866          \\
SDF-PDF                      & 21.89          & \textbf{0.885} & 20.88          & \textbf{0.869} \\
\end{tabular}
}
\end{center}
\vspace{-1.3em}
\caption{\textbf{Results of novel view synthesis of training poses and novel poses on MonoCap dataset in terms of PSNR and
SSIM (higher is better).} These methods are trained on one view and tested on ten or eleven views. The results show that our proposed methods can generate high-quality novel view and novel pose image synthesis results with even one input view.}
\vspace{-1.2em}
\label{table:monocap_result}
\end{table}

%#################################################################################################
\begin{figure*}[t]
\centering
\vspace{1em}
\includegraphics[width=1\linewidth]{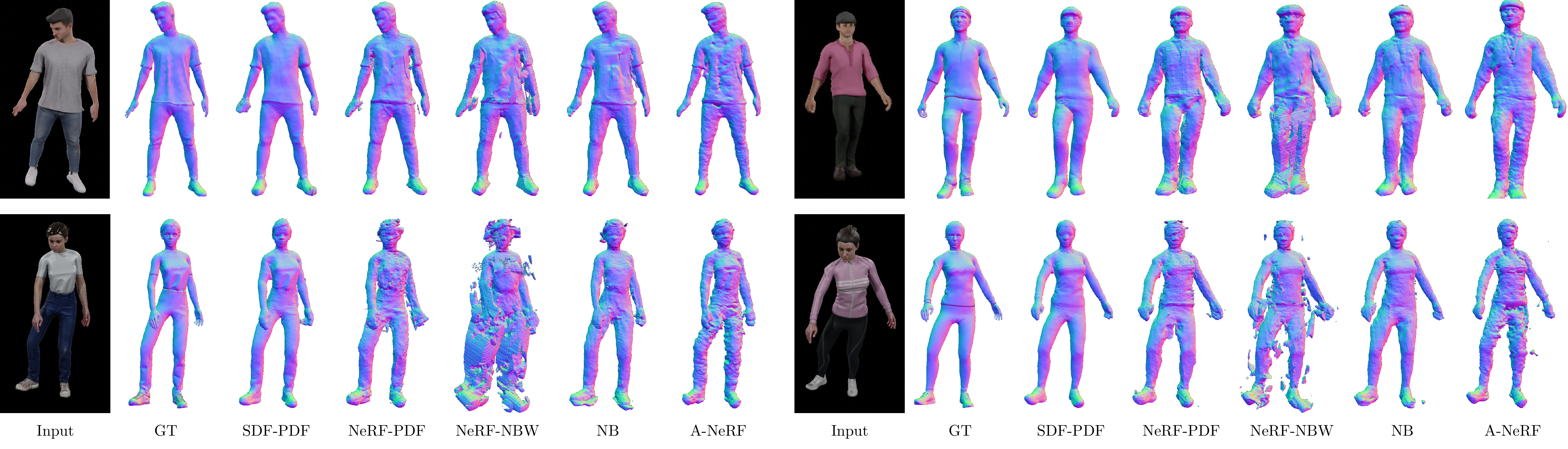}
\vspace{-1em}
\caption{\textbf{3D reconstruction on the SyntheticHuman dataset.} The results in the first row are reconstructions from 4-view videos, and the results in the second row are reconstructions from monocular videos. Our method ``SDF-PDF" significantly outperforms other methods. }
\vspace{-1em}
\label{fig:synthetichuman_reconstruction}
\end{figure*}
%#################################################################################################

\begin{table*}
\begin{center}
\scalebox{0.95}{
\tablestyle{2.5pt}{1.2}
\begin{tabular}{c|x{30}x{30}x{30}x{30}x{30}x{30}|x{30}x{30}x{30}x{30}x{30}x{30}}
& \multicolumn{6}{c|}{P2S$\downarrow$} & \multicolumn{6}{c}{CD$\downarrow$} \\[.1em]
\shline
& D-NeRF \cite{pumarola2020d} & NB \cite{peng2020neural} & A-NeRF \cite{su2021anerf} & NeRF-NBW & NeRF-PDF & SDF-PDF & D-NeRF \cite{pumarola2020d} & NB \cite{peng2020neural} & A-NeRF \cite{su2021anerf} & NeRF-NBW & NeRF-PDF & SDF-PDF \\
\hline
S1      & 3.49 & 1.44 & 1.30 & 4.30 & 1.59 & \textbf{0.75} & 2.40 & 1.39 & 1.29 & 2.94 & 1.45 & \textbf{0.86} \\
S2      & 3.38 & 1.68 & 1.39 & 4.66 & 1.74 & \textbf{0.70} & 2.45 & 1.48 & 1.22 & 3.03 & 1.51 & \textbf{0.81} \\
S3      & 3.96 & 1.52 & 1.80 & 4.45 & 1.61 & \textbf{0.62} & 2.71 & 1.42 & 1.53 & 3.02 & 1.40 & \textbf{0.81} \\
S4      & 4.18 & 1.20 & 1.46 & 2.84 & 1.58 & \textbf{0.58} & 2.85 & 1.23 & 1.28 & 2.07 & 1.40 & \textbf{0.74} \\
S5      & 1.22 & 1.20 & 37.5 & 2.87 & 1.85 & \textbf{0.66} & 1.10 & 1.14 & 36.4 & 2.33 & 1.44 & \textbf{0.65} \\
S6      & 1.76 & 1.31 & 1.17 & 2.62 & 1.97 & \textbf{0.74} & 1.43 & 1.28 & 1.07 & 2.05 & 1.54 & \textbf{0.73} \\
S7      & 1.66 & 1.61 & 1.03 & 2.88 & 2.11 & \textbf{0.69} & 1.82 & 1.74 & 1.29 & 2.59 & 2.02 & \textbf{0.65} \\
\hline
average & 2.81 & 1.42 & 6.52 & 3.52 & 1.78 & \textbf{0.70} & 2.11 & 1.38 & 6.30 & 2.57 & 1.54 & \textbf{0.75} \\
\end{tabular}
}
\end{center}
\vspace{-1em}
\caption{\textbf{Results of 3D reconstruction on SyntheticHuman dataset.} The first four rows show the results on monocular videos, and the remaining rows present the results on 4-view videos. \cite{su2021anerf} failed to converge on subject S5.}
\vspace{-1em}
\label{table:synthetichuman_result}
\end{table*}

\textbf{Results on the MonoCap dataset.} Table~\ref{table:monocap_result} summarizes the quantitative comparison of novel view synthesis between our methods and \cite{su2021anerf}, \cite{peng2020neural}, \cite{wu2020multi} under training and novel human pose. Specifically, our methods all perform better than the baselines on novel view synthesis under novel human poses. And the pose-dependent deformation field methods NeRF-PDF and SDF-PDF outperforms \cite{peng2020neural} on monocular novel view synthesis. Figure \ref{fig:novel_pose_novel_view} presents the qualitative comparison of image synthesis of our methods and baselines on MonoCap dataset, which show that our model is able to produce photo-realistic novel view synthesis under novel human poses even with only one input view.

\textbf{Results on the ZJU-MoCap dataset.} The quantitative results are listed in Table~\ref{table:zjumocap_result}. When trained on 4 camera views, our method is competitive to \cite{peng2020neural} on novel view synthesis of training poses and outperforms \cite{peng2020neural} on novel pose synthesis. Figure~\ref{fig:training_pose_novel_view} and \ref{fig:novel_pose_novel_view} present qualitative results of models that are trained on 4 camera views. When trained on the single camera view, our method significantly outperforms \cite{peng2020neural} on both training and novel poses.

\begin{table}
\begin{center}
\scalebox{0.77}{
\tablestyle{4pt}{1.2}
\begin{tabular}{l|x{25}x{25}|x{25}x{25}|x{25}x{25}|x{25}x{25}}
& \multicolumn{4}{c|}{4 Views} & \multicolumn{4}{c}{Single view} \\[.1em]
\shline
& \multicolumn{2}{c|}{Training poses} & \multicolumn{2}{c|}{Novel poses} & \multicolumn{2}{c|}{Training poses} & \multicolumn{2}{c}{Novel poses} \\[.1em]
% \shline
& PSNR$\uparrow$ & SSIM$\uparrow$ & PSNR$\uparrow$ & SSIM$\uparrow$ & PSNR$\uparrow$ & SSIM$\uparrow$ & PSNR$\uparrow$ & SSIM$\uparrow$ \\
\hline
NB \cite{peng2020neural}     & \textbf{28.15} & \textbf{0.943} & 24.05          & 0.896          & 20.90          & 0.802          & 20.32          & 0.797          \\
NeRF-NBW                     & 25.69          & 0.913          & 23.60          & 0.893          & 23.00          & 0.879          & 22.62          & 0.871          \\
NeRF-PDF                     & 27.50          & 0.935          & \textbf{24.35} & 0.904          & \textbf{23.86} & 0.894          & \textbf{22.98} & 0.883          \\
SDF-PDF                      & 27.45          & 0.940          & 24.28          & \textbf{0.909} & 23.80          & \textbf{0.901} & 22.92          & \textbf{0.889} \\
\end{tabular}
}
\end{center}
\vspace{-1.3em}
\caption{\textbf{Results of novel view synthesis of training poses and novel poses on ZJU-MoCap dataset.} These results show that all three representations of our method perform better on novel pose synthesis.}
\vspace{-1.2em}
\label{table:zjumocap_result}
\end{table}

% \vspace{-1em}

\subsection{Performance on 3D reconstruction}

To validate our method on 3D reconstruction, we compare with \cite{pumarola2020d, peng2020neural, su2021anerf}. Because \cite{wu2020multi} does not reconstruct the human geometry and \cite{liu2021neural, xu2021h} do not release the code, we do not compare with them. We use the Marching Cubes algorithm \cite{lorensen1987marching} to extract the underlying human geometry from the neural field. For methods that use the density field, we empirically set the threshold to extract the geometry when applying the Marching Cubes. For our proposed representation ``SDF-PDF", we set the threshold as zero.

\textbf{Results on the SyntheticHuman datasets.} Table~\ref{table:synthetichuman_result} compares our method with \cite{peng2020neural, peng2021animatable, su2021anerf} in terms of the P2S and CD metrics. \cite{peng2020neural, peng2021animatable, su2021anerf} and our ``NeRF-NBW", ``NeRF-PDF" representations model the human geometry with the volume density field, while our ``SDF-PDF" representation adopt the signed distance field. We empirically set the threshold of volume density to extract the geometry of \cite{peng2020neural, peng2021animatable, su2021anerf} and our ``NeRF-NBW", ``NeRF-PDF". Our representation ``SDF-PDF" significantly outperforms baseline methods\cite{peng2020neural, peng2021animatable, su2021anerf} by a margin of at least 0.72 in terms of P2S metric and 0.63 in terms of CD metric. Figure~\ref{fig:synthetichuman_reconstruction} presents the qualitative comparison.

\textbf{Results on real data.} To further validate the effectiveness of our method, we also perform reconstruction experiments on the real data. Since there are not ground-truth human geometries on real data, we only present the qualitative comparisons.

\begin{figure*}[t]
\centering
\vspace{1em}
\includegraphics[width=1\linewidth]{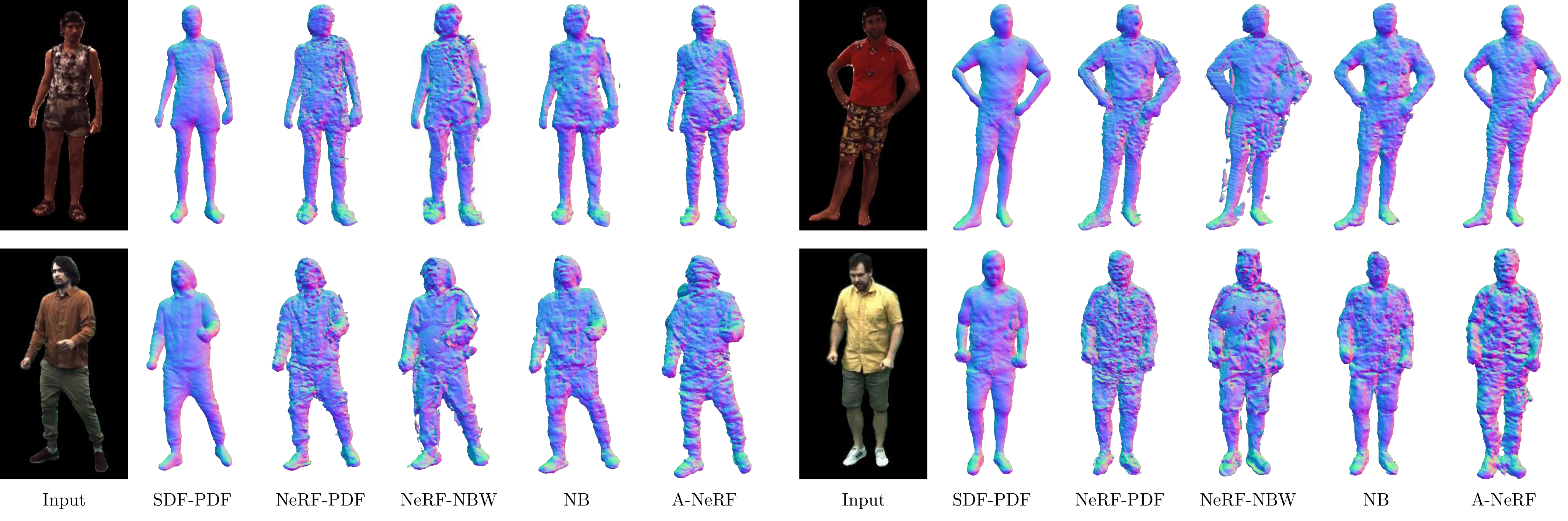}
\vspace{-1em}
\caption{\textbf{3D reconstruction on the Human3.6M and MonoCap dataset.} The results in the first row are reconstructions from 3-view videos, and the results in the second row are reconstructions from monocular videos. Our method ``SDF-PDF" significantly outperforms other methods.}
\vspace{-1em}
\label{fig:realdata_reconstruction}
\end{figure*}

Figure~\ref{fig:realdata_reconstruction} presents the reconstruction results on the Human3.6M and MonoCap dataset. The methods are trained on one view on the MonoCap dataset and three views on the Human3.6M dataset. As shown in the results, ``SDF-PDF" is able to reconstruct high-quality human geometries.

%#################################################################################################
%#################################################################################################

\subsection{Ablation studies}

\label{sec:ablation}

We conduct ablation studies to analyze how design choices and training data affect the performance of our method.

%#################################################################################################
\begin{figure}[t]
\centering
\includegraphics[width=1\linewidth]{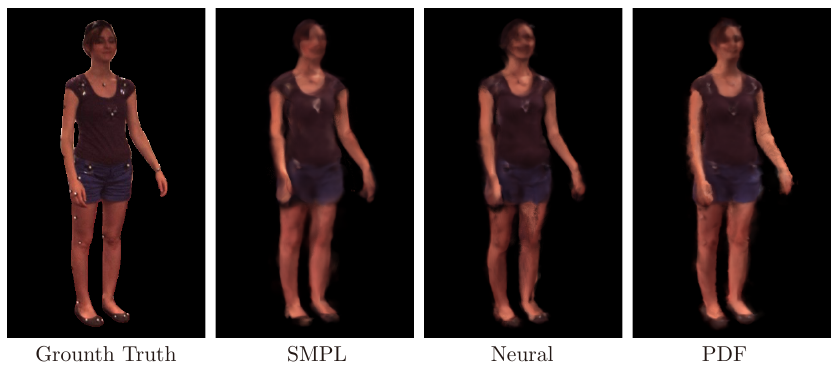}
\vspace{-1.5em}
\caption{\textbf{Qualitative comparison of different deformation fields} on subject ``S1". ``SMPL" means SMPL blend weight field, ``Neural" means neural blend weight field, and ``PDF" means pose-dependent displacement field.}
\label{fig:nsf_ssf}
\end{figure}
%#################################################################################################

\textbf{Canonical human model.} Our method proposes two ways to represent the human geometry. One is the density field, and the other one is the signed distance field. Tables \ref{table:h36m_result}, \ref{table:monocap_result}, \ref{table:zjumocap_result} show that the two representations give similar performance in terms of image synthesis on the Human3.6M, MonoCap, and ZJU-MoCap datasets. While for 3D reconstruction, signed distance field can produce better reconstruction results than density field. In Table~\ref{table:synthetichuman_result}, ``SDF-PDF" significantly outperforms ``NeRF-PDF" in terms of 3D reconstruction on the SyntheticHuman dataset. The qualitative results in Figure~\ref{fig:synthetichuman_reconstruction} and \ref{fig:realdata_reconstruction} also indicate that the reconstruction results of ``SDF-PDF" are better.

Our method uses the strategy in VolSDF \cite{yariv2021volume} to render the ``SDF-PDF". NeuS \cite{wang2021neus} proposes another way to perform volume rendering of SDF that ensures unbiased surface reconstruction based on the first-order approximation of SDF. To analyze the influence of the volume rendering scheme, we render ``SDF-PDF" with the strategy in NeuS \cite{wang2021neus}, and retrain our model on the SyntheticHuman dataset, which gives 0.79 P2S and 0.89 CD on average. ``SDF-PDF" with the rendering technique in VolSDF \cite{yariv2021volume} has a better performance, which gives 0.70 P2S and 0.75 CD on average. The reason may be that deformed signed distance fields violate the first-order approximation of SDF in \cite{wang2021neus}.

\textbf{Pose-driven deformation field.} We introduce two ways to improve the SMPL blend weight field. One is neural blend weight field, and the other one is pose-dependent displacement field. To validate the effectiveness, we train NeRF with SMPL blend weight field on the Human3.6M dataset and evaluate in terms of the novel pose synthesis performance, which gives 23.61 PSNR on average. In contrast, ``NeRF-NBW" and ``NeRF-PDF" produce 23.65 PSNR and 23.78 PSNR, respectively. Table~\ref{table:nsf_ssf} summarizes the quantitative comparisons. Figure~\ref{fig:nsf_ssf} additionally presents the qualitative results, indicating that the proposed strategies improve the performance.

We also explore which representation is better for producing the deformation field. On the Human3.6M dataset, ``NeRF-NBW" has similar performance with ``NeRF-PDF". However, on the MonoCap, ZJU-MoCap, and SyntheticHuman datasets, ``NeRF-NBW" achieves worse performance than ``NeRF-PDF" in terms of both image synthesis and 3D reconstruction. A reason may be that human poses of these three datasets are estimated from multi-view images based on the marker-less pose estimation system, which could not be very accurate. Because neural blend weight field totally models the human motion as the skeleton-driven deformation, it is more sensitive to the pose accuracy than the pose-dependent displacement field.

\begin{table}
\begin{center}
\tablestyle{4pt}{1.25}
\begin{tabular}{x{130}|x{35}|x{35}}
& PSNR & SSIM \\[.1em]
\shline
SMPL blend weight field       & 23.61             & 0.884\\
Neural blend weight field     & 23.64             & 0.890 \\
Pose-dependent displacement field & \textbf{23.78}    & \textbf{0.892} \\
\end{tabular}
\end{center}
\vspace{-1em}
\caption{\textbf{Comparison of different deformation fields} on the Human3.6M dataset.}
\vspace{-1em}
\label{table:nsf_ssf}
\end{table}

\textbf{Network architecture.} Our method adopts the network of \cite{yariv2020multiview}, while the preliminary version of this work \cite{peng2021animatable} uses the network of NeRF \cite{mildenhall2020nerf}. The newly adopted network has a bigger color head than NeRF's network. The detailed network architectures are described in the supplementary material. On the Human3.6M dataset, the original network in \cite{peng2022animatable} gives 22.55 PSNR on novel pose synthesis, while the newly adopted network gives 23.65 PSNR, indicating that the bigger network improves the rendering performance.

\begin{figure}[t]
\centering
\includegraphics[width=1\linewidth]{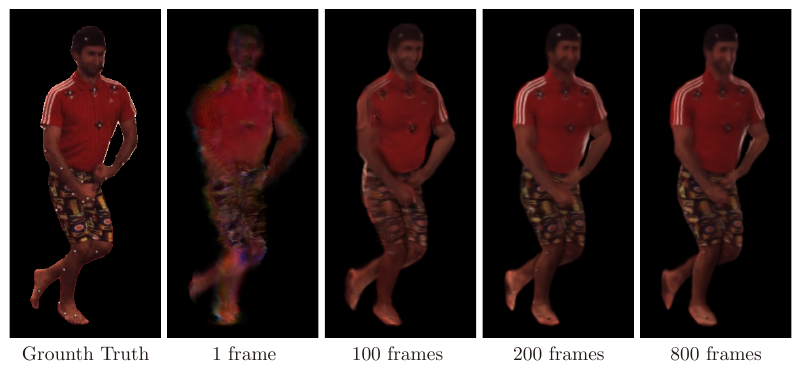}
\vspace{-1.8em}
\caption{\textbf{Comparison of models trained with different numbers of video frames} on the subject ``S9".}
\label{fig:video_length}
\end{figure}

\textbf{Number of training frames.} To explore the impact of the number of training frames, we perform ablation studies on the subject ``S9" of the Human3.6M dataset. The subject ``S9" is selected as it contains richer human motions than other subjects of Human3.6M dataset. We take 1, 100, 200 and 800 video frames for training and test the models on the same motion sequence. Table~\ref{table:video_length} lists the quantitative results of our models trained with different numbers of video frames. The results demonstrate that training on the video helps the representation learning, but the network have difficulty in fitting very long videos. Empirically, we find that 150$\sim$300 frames are suitable for most subjects. Figure~\ref{fig:video_length} presents the qualitative comparisons.

%#################################################################################################
%#################################################################################################

\begin{table}
\begin{center}
\tablestyle{4pt}{1.05}
\begin{tabular}{x{35}|x{35}|x{35}|x{35}|x{35}}
Frames & 1 & 100 & 200 & 800 \\[.1em]
\shline
PSNR & 22.46 & 24.14 & \textbf{24.68} & 24.48 \\
SSIM & 0.847 & 0.888 & \textbf{0.898} & 0.895 \\
\end{tabular}
\end{center}
\vspace{-1.5em}
\caption{\textbf{Results of models trained with different numbers of video frames} on ``S9" of Human3.6M dataset.}
% \vspace{-1.3em}
\label{table:video_length}
\end{table}

\begin{table}[!t]
\begin{center}
\tablestyle{4pt}{1.05}
\begin{tabular}{x{35}|x{35}|x{35}|x{35}}
& 1 view & 2 views & 3 views \\[.1em]
\shline
PSNR & 24.21 & 24.34          & \textbf{24.45} \\
SSIM & 0.891 & \textbf{0.898} & \textbf{0.898} \\
\end{tabular}
\end{center}
\vspace{-1.3em}
\caption{\textbf{Results of models trained with different numbers of camera views} on ``S9" of Human3.6M dataset.}
\label{table:view}
\end{table}

\begin{figure}[t]
\centering
\includegraphics[width=1\linewidth]{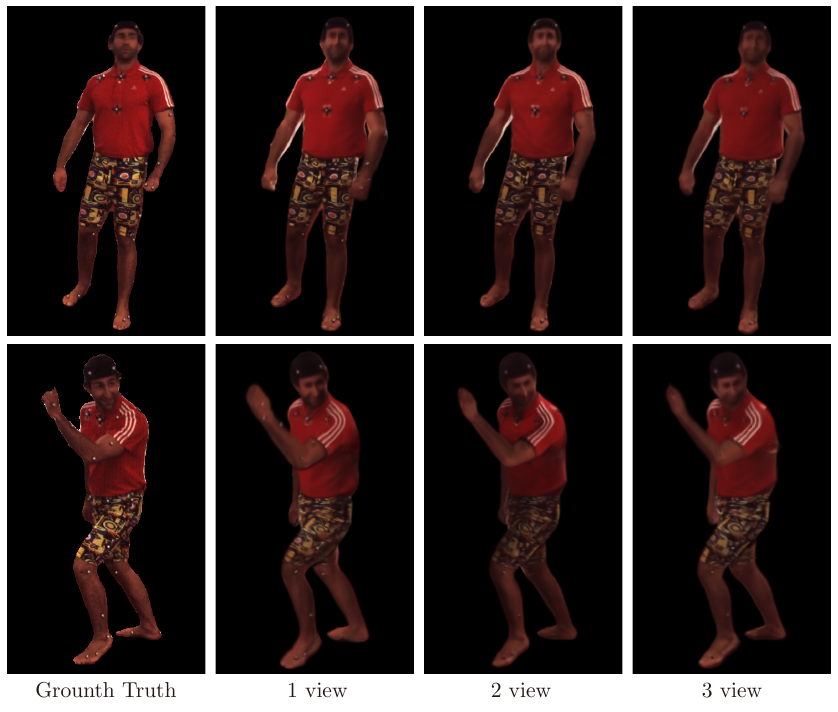}
\vspace{-1.8em}
\caption{\textbf{Comparison of models trained with different numbers of camera views} on the subject ``S9".}
\label{fig:view}
\end{figure}

\textbf{Number of input views.} To explore the impact of the number of input views, we take one view for test and select 1, 2, and 3 nearest views for training. Table~\ref{table:view} compares the performances of models trained with different numbers of input views. Surprisingly, the three models have similar quantitative performances, which in turn illustrates the effectiveness of our algorithm. Figure~\ref{fig:view} further compares the three models, which shows that the model trained on 3 views renders more accurate color. It is worth noting that the model trained on a single view already achieves decent rendering quality.

% on one subject (S9) of the H36M \cite{ionescu2013human3} dataset in terms of the novel pose synthesis performance. First, to analyze the benefit of learning $F_{\Delta \mathbf{w}}$, we compare neural blend weight field with SMPL blend weight field. Then, to explore the influence of human pose accuracy, we estimate SMPL parameters from predicted human poses \cite{cao2018openpose, joo2018total} and perform training on these parameters. Finally, we explore the performances of our method under different numbers of video frames and camera views. Tables \ref{table:nsf_ssf}, \ref{table:human_pose}, \ref{table:video_length},  and \ref{table:view} summarize the results of ablation studies.

% \subsection{Running time}

% For $512 \times 512$ images, our algorithm takes 1.09s to render an image on a desktop with an Intel i7 3.7GHz CPU and a GTX 1080 Ti GPU. Specifically, our implementation takes 0.39s for predicting the color and density fields, 0.63s for predicting the blend weight fields, and 0.07s for volume rendering. Because the number of points sampled along the ray is only 64 and the scene bound of a human is small, the rendering speed of our method is relatively fast.

%#################################################################################################
%#################################################################################################

%#################################################################################################
\begin{figure}[t]
\centering
\includegraphics[width=1\linewidth]{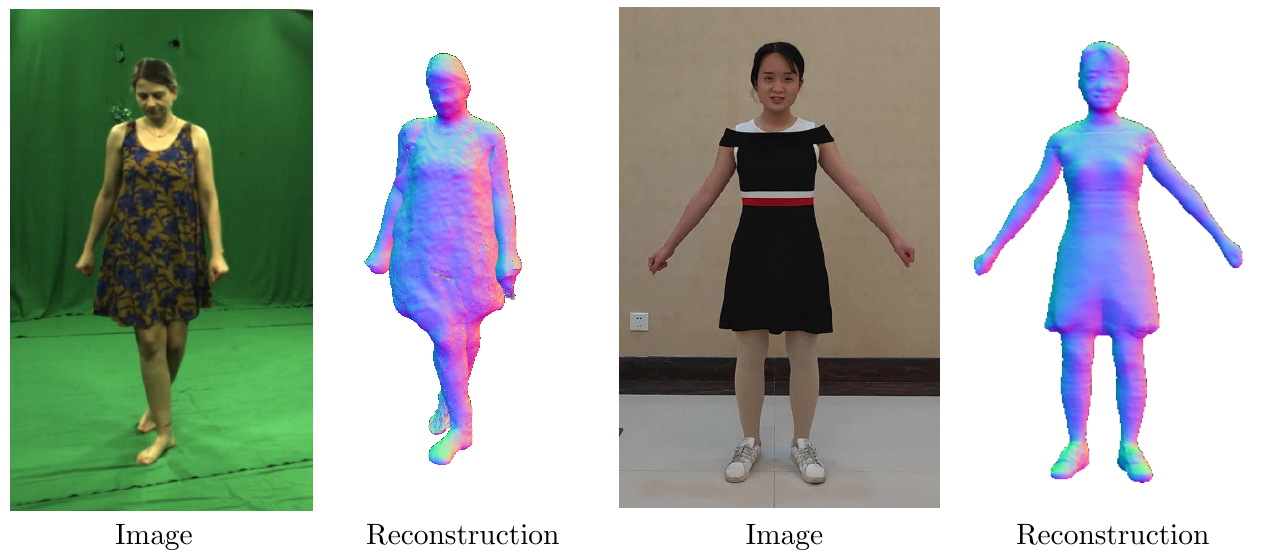}
\vspace{-1.5em}
\caption{\textbf{Reconstruction results of performers wearing loose clothes.} The images present the target performers. The reconstruction results indicate that our method is able to produce reasonable human geometries. However, we cannot reconstruct detailed cloth wrinkles in the images.}
\label{fig:loose_cloth}
\end{figure}
%#################################################################################################

\section{Limitations}

\label{section:limitations}

Combining canonical implicit neural representation with pose-driven deformation fields enables us to obtain impressive performances on creating animatable human models from videos. However, our method has a few limitations. 1) The skeleton-driven deformation model \cite{lewis2000pose} cannot express the complex non-rigid deformations of garments. As a result, the performance of our method tends to degrade when reconstructing performers that wear loose clothes. 2) Same to NeRF, our proposed model is trained per-scene, which requires a lot of time to produce animatable human models. Generalizing the networks across different videos and reducing training time is left as future work. 3) Moreoever, the rendering time of our model is a bit high. It is could be solved with recent caching-based techniques \cite{yu2021plenoctrees, hedman2021snerg}.

To explore the performance of our method on loose clothing, we collect two image sequences from \cite{habermann2020deepcap} and \cite{jiang2022selfrecon}. Both human subjects wear loose clothes. The image sequence from \cite{habermann2020deepcap} is selected from 700-th frame to 1000-th frame of the subject ``Magdalena", which performs arbitrary motions and has complex non-rigid cloth deformations. The performer from \cite{jiang2022selfrecon}, denoted as ``SelfRot", performs the self-rotation that produces small non-rigid deformations. Figure~\ref{fig:loose_cloth} presents the reconstruction results, which show that our method can reconstruct reasonable human shapes but cannot recover detailed cloth wrinkles.

\section{Conclusion}

We introduced a novel dynamic human representation for modeling animatable human characters from multi-view videos. Our method augments a canonical neural field with a pose-driven deformation field that transforms observation-space points to the canonical space. The pose-driven deformation field is constructed based on the skeleton-driven deformation framework, where we explored using neural blend weight field and pose-dependent displacement field to produce the deformation field. We showed that representing the canonical human model as the neural radiance field works well for the human modeling from videos. And replacing the density field with the signed distance field further improves the performance on 3D reconstruction. The animatable implicit neural representation is learned over the multi-view video with volume rendering. After training, our method can synthesize free-viewpoint videos of a performer given novel motion sequences. Experiments on the Human3.6M, MonoCap, ZJU-MoCap, and SyntheticHuman datasets demonstrated that the proposed model achieves state-of-the-art performances on image synthesis and 3D reconstruction.

\ifCLASSOPTIONcompsoc
  \section*{Acknowledgments}
\else
  \section*{Acknowledgment}
\fi

The authors from Zhejiang University would like to acknowledge support from NSFC (No. 62172364). 

\ifCLASSOPTIONcaptionsoff
  \newpage
\fi



%\bibliographystyle{IEEEtran}
\bibliographystyle{IEEEtran}
\bibliography{main}


\begin{IEEEbiography}{}
\end{IEEEbiography}

\begin{IEEEbiographynophoto}{}
\end{IEEEbiographynophoto}

%\newpage

\begin{IEEEbiographynophoto}{}
\end{IEEEbiographynophoto}


%\vfill


\newpage

\setcounter{section}{0}
\section*{Supplementary Material}

\textbf{Overview.} The supplementary material has the following contents:

\begin{itemize}
    \item Section~\ref{sec:exp} provides descriptions of baseline methods, datasets, and evaluation metrics.
    \item Section~\ref{sec:implementation} describes the implementation details, including the derivation of transformation matrices, network architectures, volume rendering process, and training strategy.
    \item Section~\ref{sec:discussion} provides more discussions, which aim to sufficiently evaluate our method.
\end{itemize}

\section{Experimental details}

\label{sec:exp}

\subsection{Baseline methods}

% \cite{pumarola2020d, peng2020neural, su2021anerf, wu2020multi}

We compare with state-of-the-art image synthesis methods \cite{wu2020multi, peng2020neural, pumarola2020d, su2021anerf} that also utilize SMPL priors. Same to our method, these methods train a separate network for each video. 1) NHR \cite{wu2020multi} extracts 3D features from input point clouds and renders them into 2D feature maps, which are then transformed into images using 2D CNNs. Since dense point clouds are difficult to obtain from sparse camera views, we take SMPL vertices as input point clouds. 2) Neural body \cite{peng2020neural} anchors a set of latent codes on the vertices of SMPL and uses a network to regress neural radiance fields from the latent codes, which are then rendered into images using volume rendering. 3) D-NeRF \cite{pumarola2020d} decomposes the dynamic human into a canonical human model and a deformation field. The human model is represented as a neural radiance field, and the deformation field is predicted by an MLP network that takes time index and spatial location as input. 4) A-NeRF \cite{su2021anerf} constructs the skeleton-relative embedding for input 3D points to represent the animatable human model and jointly optimizes the input skeleton poses and network parameters during training.

\subsection{Dataset details}

\textbf{Human3.6M \cite{ionescu2013human3}.} Following \cite{peng2021animatable}, we use three camera views for training and test on the remaining view. \cite{peng2021animatable} select video clips from the action ``Posing" of S1, S5, S6, S7, S8, S9, and S11. The number of training frames and test frames is described in Table~\ref{table:h36m_dataset}. 

\textbf{MonoCap \cite{peng2022animatable}.} It consists of two videos ``Lan" and ``Marc" from DeepCap dataset \cite{habermann2020deepcap}, and two videos ``Olek" and ``Vlad" from DynaCap dataset \cite{habermann2021real}. ``Lan" is selected from 620-th frame to 1220-th frame in the original video. ``Marc" is selected from 35000-th frame to 35600-th frame. ``Olek" is selected from 12300-th frame to 12900-th frame. ``Vlad" is selected from 15275-th frame to 15875-th frame. Each clip has 300 frames for training and 300 frames for evaluating novel pose synthesis, respectively. We use the 0-th camera as the training view for ``Lan" and ``Marc". The 44-th camera is selected as the training view for ``Olek". The training view of ``Vlad" is the 66-th camera. We uniformly select ten cameras from the remaining cameras for test.

\textbf{ZJU-MoCap \cite{peng2020neural}.} It records multi-view videos with 21 cameras and collects human poses using the marker-less motion capture system. Table~\ref{table:zjumocap_dataset} lists the number of training and test frames in the ZJU-MoCap dataset.

\textbf{SyntheticHuman \cite{peng2022animatable}.} It contains 7 human characters. Subjects S1, S2, S3, and S4 perform rotation with A-pose, which are rendered into monocular videos. Subjects S5, S6 and S7 perform random actions, which are rendered into 4-view videos. The number of video frames is listed in Table~\ref{table:syntheticsuman_dataset}.

\begin{table}
\begin{center}
\scalebox{0.92}{
\tablestyle{4pt}{1.05}
\begin{tabular}{c|x{20}x{20}x{20}x{20}x{20}x{20}x{20}}
\hline 
subject    & S1  & S5  & S6  & S7  & S8  & S9 & S11\\[.1em]
\shline
training    & 150 & 250 & 150 & 300 & 250 & 260 & 200 \\
test        & 49  & 127 & 83  & 200 & 87  & 133 & 82\\
\hline 
\end{tabular}
}
\end{center}
\vspace{-1em}
\caption{\textbf{The number of training frames and test frames of the Human3.6M dataset.}}
\label{table:h36m_dataset}
\end{table}

\begin{table}
\begin{center}
    \scalebox{0.75}{
\tablestyle{4pt}{1.05}
\begin{tabular}{c|x{20}x{20}x{25}x{25}x{25}x{28}x{25}x{25}x{25}}
\hline 
subject    & Twirl  & Taichi  & Swing1  & Swing2  & Swing3 & Warmup & Punch1 & Punch2 & Kick \\[.1em]
\shline 
training   & 60   & 400  & 300 & 300 & 300 & 300 & 300 & 300 & 400 \\
test       & 1000 & 1000 & 356 & 559 & 358 & 317 & 346 & 354 & 700 \\
\hline 

\end{tabular}
}
\end{center}
\vspace{-1em}
\caption{\textbf{The number of video frames for each subject in the ZJU-MoCap dataset.}}
\label{table:zjumocap_dataset}
\end{table}

\begin{table}
\begin{center}
\scalebox{0.92}{
\tablestyle{4pt}{1.05}
\begin{tabular}{c|x{20}x{20}x{20}x{20}x{20}x{20}x{20}}
\hline 
subject    & S1  & S2  & S3  & S4  & S5  & S6 & S7\\[.1em]
\shline 
training    & 69 & 300 & 70 & 100 & 100 & 100 & 70 \\
\hline 

\end{tabular}
}
\end{center}
\vspace{-1em}
\caption{\textbf{The number of video frames for each subject in the SyntheticHuman dataset.}}
\label{table:syntheticsuman_dataset}
\end{table}

\subsection{Evaluation metrics}

We follow \cite{peng2020neural} to calculate the metrics of image synthesis. Specifically, the 3D human bounding box is first projected to produce a 2D mask. Then, we calculate the PSNR metric based on the pixels inside the 2D mask. Since the SSIM metric require the image input, we compute the 2D box that bounds the 2D mask and crop the image within the box, which is used to calculate the SSIM metric. For the SyntheticHuman dataset, we calculate the reconstruction metrics every 10-th frame. For the Human3.6M and MonoCap datasets, we calculate the metrics of image synthesis every 30-th frame.

%#################################################################################################
\begin{figure}[t]
\centering
\includegraphics[width=1\linewidth]{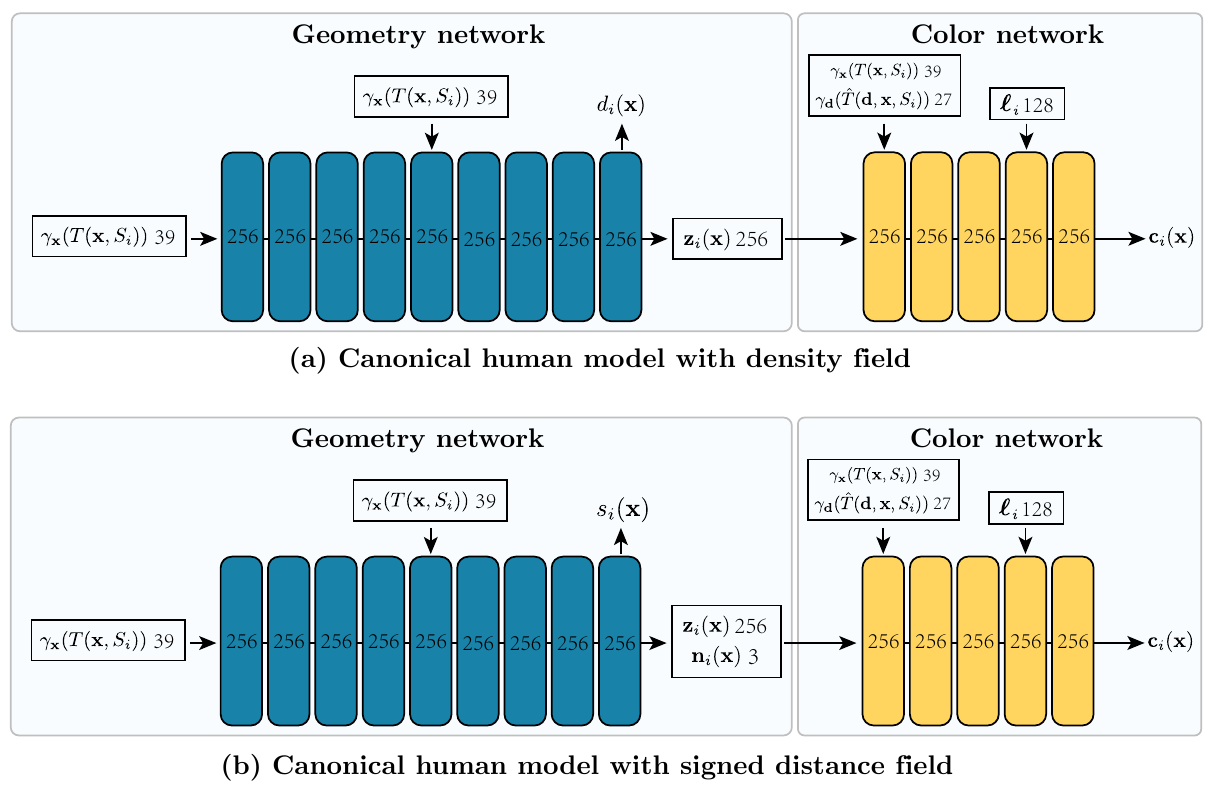}
\vspace{-1em}
\caption{\textbf{Canonical human model.} We present two types of canonical human models. (a) One models the human geometry with density field, and (b) the other one models the geometry with the signed distance field. All layers are linear layers with softplus activations except for the final layer. The dimension of the input is shown in each block.}
\label{fig:sdf}
\end{figure}
%#################################################################################################

%#################################################################################################
\begin{figure}[t]
\centering
\includegraphics[width=0.8\linewidth]{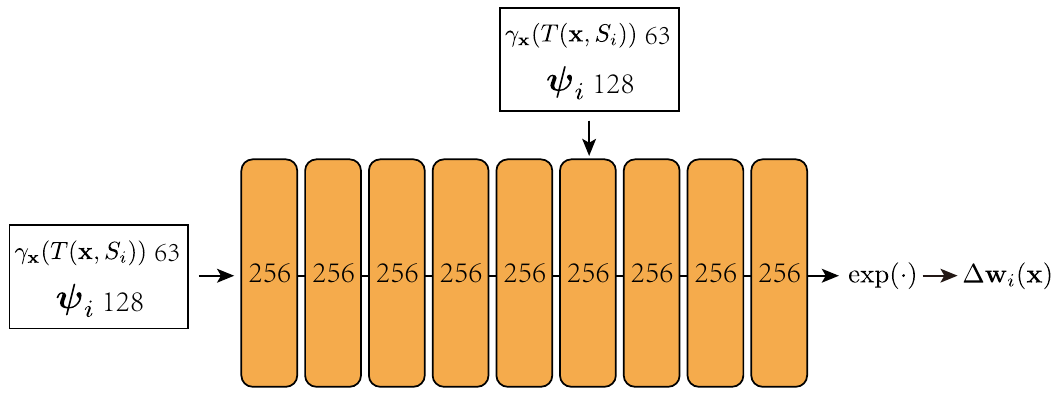}
\vspace{-1em}
\caption{\textbf{Neural blend weight field.} All layers are linear layers with ReLU activations except for the final layer. The network takes the positional encoding of spatial point $\gamma_\mathbf{x}(T(\mathbf{x}), S_i)$ and the per-frame latent code for $\psi_i$ as input.}
\vspace{-1em}
\label{fig:nbw}
\end{figure}
%#################################################################################################

%#################################################################################################
\begin{figure}[t]
\centering
\includegraphics[width=0.8\linewidth]{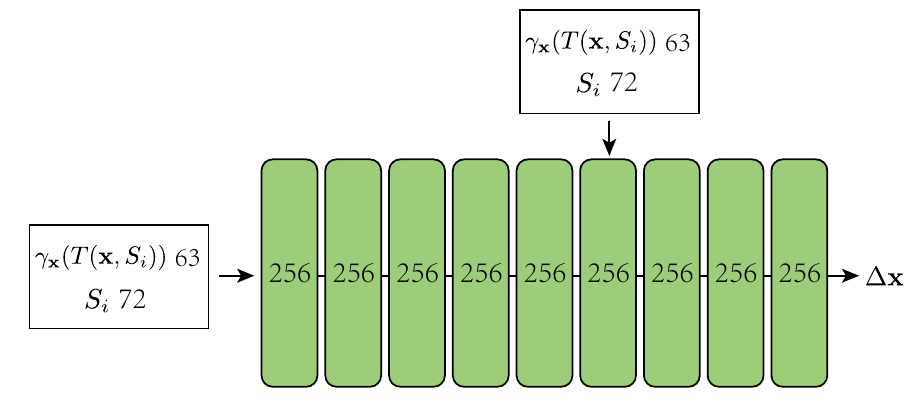}
\vspace{-1em}
\caption{\textbf{Pose-dependent displacement field.} All layers are linear layers with ReLU activations except for the final layer. The network takes the positional encoding of spatial point $\gamma_\mathbf{x}(T(\mathbf{x}), S_i)$ and the human pose $S_i$ as input.}
\vspace{-1em}
\label{fig:displacement}
\end{figure}
%#################################################################################################

\section{Implementation details}

\label{sec:implementation}

\subsection{Derivation of transformation matrices}

Given the human skeleton, the LBS model \cite{lewis2000pose} calculates the transformation matrices of body parts to produce the deformation field. We represent the human skeleton as $(\mathbf{J},\boldsymbol{\theta})$, where $\mathbf{J}\in \mathbb{R}^{K\times3}$ denotes the joint locations of $K$ joints and $\boldsymbol{\theta} \in \mathbb{R}^{3(K+1)\times1}=[\boldsymbol{\omega}_0^T,...,\boldsymbol{\omega}_K^T]$ denotes the $(K+1)$ relative rotation of body part with respect to its parent part in a kinematic tree using the axis-angle representation. Then, the transformation matrix of part $k$ from canonical pose $\boldsymbol{\theta}_c$ to target pose $\boldsymbol{\theta}_t$ can be represented as 

\begin{equation}
	\label{eq:motion_basis_f}
	G_k = A_k(\mathbf{J},\boldsymbol{\theta}_t)A_k(\mathbf{J},\boldsymbol{\theta}_c)^{-1},
\end{equation}

\begin{equation}
	\label{eq:motion_basis_g}
	A_k(\mathbf{J},\boldsymbol{\theta}) = \prod_{i \in P(k)}
	\begin{bmatrix}
		R(\boldsymbol{\omega}_i) && \mathbf{j}_i \\
		0 && 1
	\end{bmatrix},
\end{equation}
where $R(\boldsymbol{\omega}_i)\in \mathbb{R}^{3\times3}$ is the converted rotation matrix of $\boldsymbol{\omega}_i$ via the Rodrigues formula, $\mathbf{j}_i$ is the $i$-th joint center, and $P(k)$ is the ordered set of parent joints of joint $k$. In practice, we adopt the SMPL skeleton \cite{loper2015smpl}, which has $K=24$ parts, but this idea applies to other human skeletons \cite{cao2018openpose, ionescu2013human3}.

\subsection{Network architectures}

Figures \ref{fig:sdf}, \ref{fig:nbw} and \ref{fig:displacement} illustrate network architectures of canonical human model, neural blend weight field $F_{\Delta \mathbf{w}}$, and pose-dependent displacement field $F_{\Delta \mathbf{x}}$, respectively. We perform positional encoding \cite{mildenhall2020nerf} to the spatial point and viewing direction. For the canonical human model, 6 frequencies are used when encoding spatial position, and 4 frequencies are used when encoding viewing direction. For the blend weight field and displacement field, 10 frequencies are used when encoding spatial position. The dimension of appearance code $\boldsymbol{\ell}_i$ is 128.

The color network $F_{\mathbf{c}}$ takes the canonical-space viewing direction as input to better approximate the radiance function. To obtain the canonical-space viewing direction, we transform the observation-space viewing direction $\mathbf{d}$ to the canonical space based on the LBS model. Denote the weighted sum of transformation matrices in the LBS model as $[R^{*}_i(\mathbf{x}); t_i(\mathbf{x})] = \sum w^k_i(\mathbf{x}) G^k_i$. The deformation $\hat{T}(\mathbf{d}, \mathbf{x}, S_i)$ that transforms the viewing direction to the canonical space is defined as:
\begin{equation}
    \hat{T}(\mathbf{d}, \mathbf{x}, S_i) = R^{*}_i(\mathbf{x})\mathbf{d},
\end{equation}
where $R^{*}_i(\mathbf{x})$ is a $3 \times 3$ matrix. To validate the benefit of using the canonical-space viewing direction, we evaluate our model with the world-space viewing direction on the subject ``S9" of Human3.6M dataset, which gives 23.65 PSNR and 0.887 SSIM on novel pose synthesis. In contrast, our model with the canonical-space viewing direction gives 24.45 PSNR and 0.898 SSIM, indicating that using the canonical-space viewing improves the performance.

\subsection{Volume rendering}

% 1. 用volume rendering渲染
% 2. 发射每一条ray, 算bounds
% 3. 预测每个点的density和color
% 4. 渲染公式
% 5. SDF convert to density

We can use volume rendering techniques \cite{kajiya1986rendering, mildenhall2020nerf} to render the animatable implicit neural representation from particular viewpoints. Given a pixel at frame $i$, we emit the camera ray and calculate the near and far bounds by intersecting the camera ray with the 3D bounding box of the SMPL model. Then, we use a stratified sampling approach \cite{mildenhall2020nerf} to sample $N_k$ points $\{\mathbf{x}_k\}^{N_k}_{k=1}$. The number of sampled points $N_k$ is set as 64 in all experiments. These points are fed into the proposed pipeline to predict the densities $\sigma_i(\mathbf{x}_k)$ and colors $\mathbf{c}_i(\mathbf{x}_k)$, which are accumulated into the pixel color $\tilde{\mathbf{C}}_i(\mathbf{r})$ using the numerical quadrature:
\begin{equation}
    \tilde{\mathbf{C}}_i(\mathbf{r}) = \sum_{k = 1}^{N_k} \alpha_i(\mathbf{x}_k) \prod_{j < k} \bigl(1 - \alpha_i(\mathbf{x}_j)) \mathbf{c}_i(\mathbf{x}_k),
\end{equation}
where $\alpha_i(\mathbf{x}_k) = 1 - \exp(-\sigma_i(\mathbf{x}_k) \delta_k)$, and $\delta_k$ is the distance between adjacent sampled points $||\mathbf{x}_{k + 1} - \mathbf{x}_{k} ||_2$.

When the human geometry is represented by the signed distance field, we first convert the predicted signed distances into volume densities and then perform the volume rendering, as \cite{yariv2021volume, wang2021neus} do. Following \cite{yariv2021volume}, we convert signed distance $s_i(\mathbf{x}_k)$ into volume density using
\begin{equation}
\sigma_i(\mathbf{x}) = 
    \begin{cases}  
        \frac{1}{\beta} \parr{1-\frac{1}{2}\exp\parr{\frac{s_i(\mathbf{x})}{\beta}}} & \text{if } s_i(\mathbf{x}) < 0, \\
        \frac{1}{2 \beta} \exp\parr{-\frac{s_i(\mathbf{x})}{\beta}} & \text{if } s_i(\mathbf{x}) \geq 0, \\
    \end{cases}
\end{equation}
where $\beta$ is a learnable parameter.

\subsection{Losses functions}

% 1. 对于不同的animatable implicit representations, 我们用不同的loss组合。
% 2. 在methods里写了L_1, L_2等loss。
% 3. 当训练NeRF-NBW的时候，用这个loss。
% 4. 当训练NeRF-PDF的时候，用这个loss。
% 5. 当训练SDF-PDF的时候，用这个loss。

In experiments, we evaluate three types of animatable implicit neural presentations, including NeRF-NBW, NeRF-PDF, and SDF-PDF, which are optimized based on different loss functions. For NeRF-NBW, the combination of the rendering loss $L_{\text{rgb}}$ and consistency loss $L_{\text{nsf}}$ is used for training, which is defined as:
\begin{equation}
    L_{\text{NeRF-NBW}} = L_{\text{rgb}} + L_{\text{nsf}}.
\end{equation}
For NeRF-PDF, we use the combination of the rendering loss $L_{\text{rgb}}$ and regularization term $L_{\Delta \mathbf{x}}$, which is defined as:
\begin{equation}
    L_{\text{NeRF-PDF}} = L_{\text{rgb}} + 0.01 L_{\Delta \mathbf{x}}.
\end{equation}
For SDF-PDF, we use the combination of the rendering loss $L_{\text{rgb}}$, mask loss $L_{\text{mask}}$, Eikonal term $L_{\text{E}}$, and regularization term $L_{\Delta \mathbf{x}}$, which is defined as:
\begin{equation}
    L_{\text{SDF-PDF}} = L_{\text{rgb}} + L_{\text{mask}} + 0.01 L_{\text{E}} + 0.01 L_{\Delta \mathbf{x}}.
\end{equation}

\subsection{Training}

% 1. In all training processes, 我们都用了Adam optimizer，initial learning rate都是1e-5, schedular都是exponential decay。
% 2. 对于pose-dependent displacement field，我们只训练一个stage。
% 3. 对于neural blend weight field，我们训练两个stage。
% 4. 在novel pose上，我们用的是一个独立的网络。

In all experiments, we use the Adam optimizer for the training, and the learning rate starts from $5e^{-4}$ and decays exponentially to $5e^{-5}$ along the optimization. Animatable implicit neural representations with the pose-dependent displacement field requires a single stage training on the input video, while the neural blend weight field requires the additional optimization on novel human poses based on the loss function $L_{\text{new}}$, which is described in the Section 3.4 of the main paper. To improve the capacity of our model, the neural blend weight field $\mathbf{w}^{\text{new}}$ of novel human poses does not share network parameters with the blend weight field $\mathbf{w}^{\text{can}}$ of the canonical human pose.

\section{Discussions}

\label{sec:discussion}

% 1. Blend weight field和displacement field一起学会怎样。-> 对比training pose和novel pose的novel view synthesis的效果。-> 如果用了neural blend weight field，会导致在novel pose上还要重新优化。
% 2. What is the performance of pose-dependent neural blend weight field.
% 3. The comparison of the amount of network parameters.
% 4. Running time analysis.

We provide more discussions on possible design choices and interesting experiments, aiming to show more insights.

\textbf{Combining neural blend weight field with pose-dependent displacement field.} The neural blend weight field can be used together with the pose-dependent displacement field to produce the deformation field. Given a human pose $S_i$ and a 3D point $\mathbf{x}$ in the observation space, we first compute the neural blend weight using $\mathbf{w}_i(\mathbf{x})$ and then leverage the LBS model to transform the observation-space point to the canonical space, resulting in the transformed point $\mathbf{x}'$. Then, the pose-dependent displacement field $F_{\Delta \mathbf{x}}$ takes $\mathbf{x}'$ as input and output the displacement to deform this point. The final point is fed into the canonical human model to predict the geometry and color. Here we use neural radiance field to represent the canonical human model. Experiments on the subject ``S9" of the Human3.6M dataset show that this strategy does not perform as well as NeRF-PDF, which gives 25.94 PSNR and 0.911 SSIM on training poses, while NeRF-PDF gives 26.03 PSNR and 0.917 SSIM. The reason is that the articulated motions could be modeled by the displacement field, leading to the local minima, as discussed in \cite{weng2022humannerf}. A possible solution is using the coarse-to-fine optimization strategy \cite{weng2022humannerf}.

\textbf{Performance of pose-dependent blend weight field.} We define an MLP network that maps the 3D point and the human pose to the residual vector of blend weight, denoted as $F'_{\Delta \mathbf{w}}: (\mathbf{x}, S) \rightarrow \Delta \mathbf{w}$. Then the residual vector of blend weight is used to update the SMPL blend weight based on
\begin{equation}
    \mathbf{w}(\mathbf{x}, S) = \text{norm}(F'_{\Delta \mathbf{w}}(\mathbf{x}, S) + \mathbf{w}^{\text{s}}(\mathbf{x}, S)).
\end{equation}
We combine this deformation field with canonical neural radiance field to represent the dynamic human. On the subject ``S9" of the Human3.6M dataset, this representation gives 0.877 PSNR on novel poses. In contrast, NeRF-NBW gives 0.885 PSNR, indicating that pose-dependent blend weight field does not generalize well to novel human poses.

\textbf{Comparison of the number of network parameters.} Table~\ref{tab:num_net_param} compares the number of network parameters of our and other methods. Our method has a smaller model size than NHR \cite{wu2020multi} and Neural Body \cite{peng2020neural}.

\begin{table}
\begin{center}
\scalebox{0.9}{
\begin{tabular}{c|x{15}x{30}x{10}x{30}x{18}x{18}x{18}}
 & NHR \cite{wu2020multi} & D-NeRF \cite{pumarola2020d} & NB \cite{peng2020neural} & A-NeRF \cite{su2021anerf} & NeRF-NBW & NeRF-PDF & SDF-PDF \\[.1em]
\shline
Params. & 18.68 & 1.21 & 4.34 & 1.78 & 1.41 & 1.38 & 1.38 \\
\end{tabular}
} % scalebox
\end{center}
\vspace{-1em}
\caption{\textbf{Number of network parameters.} Our model has fewer parameters than \cite{wu2020multi, peng2020neural}. The unit is in million.}
\label{tab:num_net_param}
\end{table}

\begin{table}
\begin{center}
\scalebox{0.9}{
\begin{tabular}{c|x{45}x{45}x{45}}
 & NeRF-NBW & NeRF-PDF & SDF-PDF \\[.1em]
\shline
Deformation           & 0.50 & 0.64 & 0.64 \\
Canonical human model & 1.11 & 1.05 & 2.38 \\
Volume rendering      & 0.02 & 0.02 & 0.02 \\
\hline
Total & 1.63 & 1.71 & 3.04
\end{tabular}
} % scalebox
\end{center}
\vspace{-1em}
\caption{\textbf{Running time.} This table analyzes the running time of three representations. The unit is in second. ``Deformation" means predicting the pose-driven deformation field and transformating the observation-space points to the canonical space. ``Canonical human model" means predicting the geometry and appearance of the canonical human model. ``Volume rendering" means accumulating the predicted geometry and appearance into pixel colors. Because SDF-PDF needs to additionally calculate the poinr normal, it takes more time to predict the geometry and color.}
\label{tab:running_time}
\end{table}

\textbf{Running time analysis.} We test the running time of NeRF-NBW, NeRF-PDF, and SDF-PDF that render a $512 \times 512$ image on a desktop with an Intel i7 3.7GHz CPU and a GTX 2080 Ti GPU. Table~\ref{tab:running_time} lists the results of running time. Because the number of points sampled along the ray is only 64 and the scene bound of a human is small, the rendering speed of our method is relatively fast.

\end{document}